\documentclass[manuscript,screen, acmlarge]{acmart}
\AtBeginDocument{%
  \providecommand\BibTeX{{%
    \normalfont B\kern-0.5em{\scshape i\kern-0.25em b}\kern-0.8em\TeX}}}

\setcopyright{acmlicensed}
\acmJournal{TKDD}
\acmYear{2024} \acmVolume{1} \acmNumber{1} \acmArticle{1} \acmMonth{1}\acmDOI{10.1145/3679018}
\usepackage{multirow}
\usepackage{graphicx}
\usepackage[ruled,linesnumbered]{algorithm2e}
\usepackage{subfigure}
\usepackage{verbatim}
\usepackage{amsmath}
\usepackage{caption}
\usepackage{ragged2e}
\usepackage{hyperref}
\begin{document}

%%
%% The "title" command has an optional parameter,
%% allowing the author to define a "short title" to be used in page headers.
\title{Meta-GPS++: Enhancing Graph Meta-Learning with Contrastive Learning and Self-Training}

%%
%% The "author" command and its associated commands are used to define
%% the authors and their affiliations.
%% Of note is the shared affiliation of the first two authors, and the
%% "authornote" and "authornotemark" commands
%% used to denote shared contribution to the research.
\author{Yonghao Liu}
%\authornote{Both authors contributed equally to this research.}
\email{yonghao20@mails.jlu.edu.cn}
\author{Mengyu Li}
%\authornotemark[1]

\email{mengyul21@mails.jlu.edu.cn}
% \affiliation{%
%   \institution{Key Laboratory of Symbolic Computation and Knowledge Engineering of the Ministry of Education, College of Computer Science and Technology, Jilin University}
%   \city{Changchun}
%   \country{China}}

\author{Ximing Li}
\email{liximing86@gmail.com}
\author{Lan Huang}
\email{huanglan@jlu.edu.cn}
\affiliation{%
  \institution{Key Laboratory of Symbolic Computation and Knowledge Engineering of the Ministry of Education, College of Computer Science and Technology, Jilin University}
  \city{Changchun}
  \country{China}}

\author{Fausto Giunchiglia}
\email{fausto.giunchiglia@unitn.it}
\affiliation{%
  \institution{University of Trento}
  \city{Trento}
  \country{Italy}
}

\author{Yanchun Liang}
\email{ycliang@jlu.edu.cn}
\affiliation{
  \institution{Zhuhai Laboratory of the Key Laboratory of Symbolic Computation and Knowledge Engineering of the Ministry of Education, Zhuhai College of Science and Technology}
  \city{Zhuhai}
  \country{China}
}

\author{Xiaoyue Feng}
\authornote{Corresponding author}
\email{fengxy@jlu.edu.cn}
\author{Renchu Guan}
\authornotemark[1]
\email{guanrenchu@jlu.edu.cn}
\affiliation{%
  \institution{Key Laboratory of Symbolic Computation and Knowledge Engineering of the Ministry of Education, College of Computer Science and Technology, Jilin University}
  \city{Changchun}
  \country{China}}

%%
%% By default, the full list of authors will be used in the page
%% headers. Often, this list is too long, and will overlap
%% other information printed in the page headers. This command allows
%% the author to define a more concise list
%% of authors' names for this purpose.
\renewcommand{\shortauthors}{Yonghao Liu and Mengyu Li et al.}
%\renewcommand{\shortauthors}{Yonghao Liu et al.}
%%
%% The abstract is a short summary of the work to be presented in the
%% article.
\begin{abstract}
Node classification is an essential problem in graph learning. However, many models typically obtain unsatisfactory performance when applied to few-shot scenarios. Some studies have attempted to combine meta-learning with graph neural networks to solve few-shot node classification on graphs. Despite their promising performance, some limitations remain. First, they employ the node encoding mechanism of homophilic graphs to learn node embeddings, even in heterophilic graphs. Second, existing models based on meta-learning ignore the interference of randomness in the learning process. Third, they are trained using only limited labeled nodes within the specific task, without explicitly utilizing numerous unlabeled nodes. Finally, they treat almost all sampled tasks equally without customizing them for their uniqueness. To address these issues, we propose a novel framework for few-shot node classification called \textbf{Meta-GPS++}. Specifically, we first adopt an efficient method to learn discriminative node representations on homophilic and heterophilic graphs. Then, we leverage a prototype-based approach to initialize parameters and contrastive learning for regularizing the distribution of node embeddings. Moreover, we apply self-training to extract valuable information from unlabeled nodes. Additionally, we adopt S$^2$ (scaling \& shifting) transformation to learn transferable knowledge from diverse tasks. The results on real-world datasets show the superiority of Meta-GPS++. Our code is available \href{https://github.com/KEAML-JLU/Meta-GPS-Plus}{\textcolor{red}{here}}.
\end{abstract}

%%
%% The code below is generated by the tool at http://dl.acm.org/ccs.cfm.
%% Please copy and paste the code instead of the example below.
%%


%%
%% Keywords. The author(s) should pick words that accurately describe
%% the work being presented. Separate the keywords with commas.

% \received{20 February 2007}
% \received[revised]{12 March 2009}
% \received[accepted]{5 June 2009}

%%
%% This command processes the author and affiliation and title
%% information and builds the first part of the formatted document.
\maketitle

\section{Introduction}
\label{introduction}
Graph learning has become a highly promising research direction in the field of deep learning due to its powerful representation capabilities for complex systems and networks in the real world. It can extract the desired node topology and feature information from graph-structured data for further data analysis. Also, it can be applied to various scenarios, such as social networks \cite{qi2011exploring, yuan2013latent}, anomaly detection \cite{dou2020enhancing, ding2020inductive}, and text classification \cite{yao2019graph, liu2021deep}. As a crucial analytical task in graph learning, node classification has gained increasing attention from the academic community. Previous pioneering works typically utilize graph neural networks (GNNs) to learn the effective node embeddings of graphs via sophisticated algorithms and have achieved satisfactory performance in many graph-related tasks \cite{klicpera2018predict, klicpera2019diffusion}. % ??

However, these approaches are commonly data-hungry, requiring substantial labeled data to obtain desirable classification performance. In practical scenarios, only a small subset of nodes in graphs are labeled \cite{ding2020graph}. For example, in social networks analysis, most users are unlabeled, while only a few users are labeled as specific categories, such as reading enthusiasts or music enthusiasts. These methods are highly prone to fall into overfitting when applied to the few-shot scenarios described above \cite{zhou2019meta}. Moreover, human labeling is somewhat impractical, as it is time-consuming and labor-intensive and may even require specialized domain knowledge. Therefore, research on few-shot learning (FSL) has garnered significant attention in academia, as it demonstrates promising performance in classification with limited labeled data, and its experimental design is more in line with real-world scenarios. Recently, many meta-learning methods have been proposed to tackle the problem of FSL for Euclidean data, including those in text \cite{kaiser2017learning, joshi2018extending} and image \cite{garcia2017few, khodadadeh2018unsupervised} domains. Drawing inspiration from the aforementioned meta-learning approaches, a plethora of methods for graph meta-learning have emerged as potential solutions to the FSL problem of graph-structured data \cite{ding2020graph, huang2020graph, liu2021relative}.

Despite some remarkable success, there are still several limitations of previous graph meta-learning methods for few-shot node classification,  which significantly constrain their ability to learn valuable information from graphs. 

(I) The graph encoding component of existing methods has been developed under the assumption that graphs are homophilic, which means that nodes with the same label are more likely to be connected \cite{pei2020geom}. Nevertheless, in reality, this assumption is frequently not applicable since many graphs exhibit heterophilic properties, which is known as the phenomenon of opposites attracting \cite{zheng2022graph}. This means that nodes with different classes may have a higher probability of being connected. For example, fraudsters prefer to connect with benign users rather than other fraudsters in online transaction networks \cite{dou2020enhancing}, and various kinds of amino acids tend to be linked to form proteins \cite{zhu2020beyond}. If the mainstream message passing mechanism in homophilic graphs is applied directly to heterophilic graphs, allowing the connected nodes to have similar features, the model will learn uninformative node representations. 

\begin{figure}
    \centering
    \includegraphics[width=0.9\textwidth]{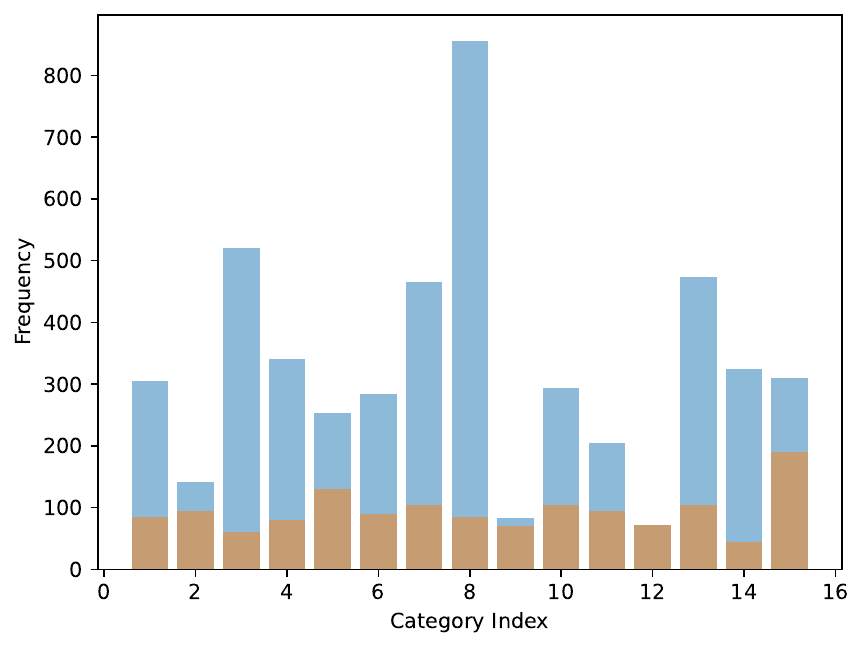}
    \caption{Distribution of the original and sampled data from the test set of the CoraFull dataset.}
    \label{distribution}
\end{figure}

(II) Several models have been developed for few-shot node classification on graphs, such as Meta-GNN \cite{zhou2019meta} and G-Meta \cite{huang2020graph}. These models utilize a popular meta-learning algorithm called model-agnostic meta-learning (MAML) to improve their performance on this task. %Several models designed for few-shot node classification on graphs, such as Meta-GNN \cite{zhou2019meta} and G-Meta \cite{huang2020graph}, utilize the classical meta-learning algorithm called model-agnostic meta-learning (MAML) \cite{finn2017model}, demonstrating its effectiveness. 
However, the stochasticity involved in the learning process of MAML may interfere with the model training and make it difficult to converge to an optimal solution. Such difficulty can be interpreted in two ways. \textit{On the one hand}, conventional MAML uses random and class-independent initialization parameters for all classes. By sharing initialization between tasks containing different classes of sampled nodes, MAML heavily depends on instance-based statistics, making it vulnerable to data noise in few-shot environments. \textit{On the other hand}, the stochasticity involved in MAML also includes the randomness in task sampling, \textit{i.e.}, random sampling instances from different classes to form tasks. Concretely, in classical few-shot learning, each task contains a support set and a query set, which are both randomly sampled. The sampled data distribution may differ significantly from the original one and may not represent the global distribution, especially when dealing with diverse and complex data distributions. To support our claim, we conduct data statistics on the entire test set of the real-world citation network CoraFull \cite{bojchevski2017deep} and 50 randomly sampled 5-way 5-shot meta-tasks. The results, as shown in Fig. \ref{distribution}, reveal a significant distribution deviation between the original and sampled data.

(III) Meta-learners in few-shot node classification often suffer from overfitting problems due to the biased distribution formed by the few labeled training examples of each task \cite{rajendran2020meta}. Previous models only focus on the selected nodes in each task and simply discard other nodes in the graph. However, these unselected nodes, although unlabeled in the specific task, may still contain rich self-supervised and task-relevant information \cite{wuinformation}. Therefore, effective utilization of such information can regularize model training, alleviate the overfitting problem, and improve the model's learning performance.

(IV) Many existing methods treat all tasks equally, disregarding the fact that tasks can exhibit significant differences in complexity and diversity. For instance, protein networks often exhibit distinct structural patterns for sequences of different types of amino acids, such as alpha helix structures and beta sheet structures \cite{dill2012protein}. To fully exploit such inter-task variability and improve model performance, it is beneficial to customize the learnable parameters for each task.

To address the aforementioned issues, we propose a novel model called Meta-GPS++, for few-shot node classification on graphs. Specifically, Meta-GPS++ consists of five essential components, including \textit{a graph network encoder}, \textit{prototype-based parameter initialization}, \textit{contrastive learning for task randomness}, \textit{self-training for model regularization}, and \textit{S$^2$ transformation for different tasks}. We have overcome the limitation of relying on the assumption of homophilic graphs and designed a method to learn informative node representations even on heterophilic graphs. Next, we introduce prototype-based initialization based on MAML to solve the problems associated with random initialization of model parameters. Then, we utilize contrastive learning to regularize the learned node embeddings and mitigate the negative impact caused by task randomness. Moreover, to fully utilize the unlabeled nodes that are not selected for the task, we introduce a self-training approach to extract the valuable self-supervised and task-relevant information from them. Finally, to adapt inter-task differences, we employ two networks to produce scaling and shifting vectors that dynamically adjust prior parameters for each new task. %(or activate different properties to adapt diverse tasks)
To summarize, the contributions of our work are as follows:

(1) We propose a novel model called Meta-GPS++ for few-shot node classification, which utilizes highly transferable meta-knowledge acquired from meta-training tasks to facilitate rapid adaptation to new tasks.

(2) We introduce several key components to address the aforementioned limitations. To our knowledge, we are the first to remove the assumption of homophilic graphs in few-shot node classification and explicitly consider the interference of stochasticity on model training in the meta-learning process. To this end, we design a corresponding graph encoder and use prototype-based parameter initialization and contrastive learning to alleviate task randomness. 

(3) We attempt to utilize out-of-task unlabeled nodes and capture differences between different tasks. To achieve this, we leverage self-training and S$^2$ transformation techniques to enhance the model's learning ability.

(4) We perform adequate experiments using a series of few-shot settings on real-world datasets. The experimental results show that the proposed Meta-GPS++ model consistently outperforms other competitive models.

Our previous work has been published at the 45th International ACM SIGIR Conference on Research and Development in Information Retrieval \cite{liu2022few}. Building upon the conference version of the paper, we substantially expand the original work in the following areas: (I) We introduce contrastive learning for the task randomness module, which encourages the model to learn the invariant node embeddings in the task and alleviates the bias caused by the task randomness in the MAML learning process. In addition, we propose using self-training for regularization, which aims to extract valuable information from previously unselected fruitful unlabeled nodes in graphs. According to the ablation experiments, removing the two modules leads to a significant performance decline in the model, which clearly demonstrates these importance. (II) To validate the superiority of our proposed Meta-GPS++, we considerably enrich the experiments. Specifically, we compare our model with several recently proposed competitive models in addition to the  originally selected baseline models, while evaluating their robustness to noisy data. Additionally, compared to the original work, we conduct more sensitivity analyses of hyperparameters to thoroughly investigate the properties of the model. (III) We perform a comprehensive review of related works and revise the flowchart to enhance the understanding of our model. Furthermore, we diligently revise the article to enhance its readability. %Furthermore, we meticulously edit the article for improved readability. 
\section{Related Work}
This section provides a review of the relevant work on GNNs, FSL, FSL on graph, and contrastive learning.

\subsection{Graph Neural Networks} 
Graphs are ubiquitous in the real world, and graph learning has been highly esteemed in academia due to its ability to model complex data. GNNs are representative graph learning methods that have developed rapidly in recent years by leveraging the powerful representation learning capabilities of neural networks \cite{wu2020comprehensive}. A series of GNN models, such as graph convolutional networks (GCNs) \cite{kipf2016semi}, graph attention networks (GATs) \cite{velivckovic2017graph}, and GraphSAGE \cite{hamilton2017inductive}, have been developed to address various practical problems
 \cite{wang2019kgat, he2020lightgcn, liu2021vplag, liu2021deep, liulocal, liu2023time, ji2023retraining, liu2024resolving}. %Graph Convolutional Network (GCN) \cite{kipf2016semi} is a pioneering work that designs a local spectral graph convolutional layer to learn node embeddings. Several GCN-based variants, such as Simplifying Graph Convolution (SGC) \cite{wu2019simplifying}, achieve better performance in semi-supervised node classification by modifying some components in the network. GraphSAGE \cite{hamilton2017inductive} generates nodes' embeddings by learning a aggregator function that samples and aggregates features from nodes' local neighborhood. Graph Attention Network (GAT) \cite{velivckovic2017graph} assigns different weights to different neighbors of a node to learn its representations by introducing self-attention mechanisms. 
However, although these models have achieved excellent performance on homophilic graphs, their performance on heterophilic graphs is unsatisfactory \cite{zheng2022graph}. Therefore, some research has been dedicated to investigating the application of these models to heterophilic graphs \cite{zhu2020beyond, chien2020adaptive, gao2023addressing}. For example, H$_{2}$GCN \cite{zhu2020beyond} promotes node representation learning on heterophilic graphs by introducing the designs of higher-order neighborhoods and combinations of intermediate representations. % and prove theoretically that the GCN layer implemented as $\mathbf{AXW}$ is more robust to perturbation than that implemented as $\mathbf{(A+I)XW}$. 
The GPR-GNN model \cite{chien2020adaptive} combines the generalized PageRank algorithm with adaptive learning of the adjacency matrix coefficients to simulate various graph filters, enabling high-quality node representations on heterophilic graphs.
However, the GNN-based models mentioned above are not capable of dealing directly with novel classes that have only a few labeled instances, \textit{i.e.}, the few-shot scenario.

\subsection{Few-shot Learning} %Since labeling data is labor-intensive, few-shot learning (FSL) is becoming increasingly popular among researchers. 
FSL aims to quickly adapt to new tasks with a few labeled instances by utilizing meta-knowledge learned from diverse meta-training tasks, which can enhance its generalization ability. Meta-learning, also known as ``learning to learn'', is a popular approach to address FSL problems. Basically, according to the framework of meta-learning algorithms, FSL methods can be mainly classified into three categories: \textit{model-based} \cite{santoro2016meta, cai2018memory, zhu2018compound}, \textit{metric-based} \cite{vinyals2016matching, sung2018learning, snell2017prototypical}, and \textit{optimization-based} \cite{finn2017model, li2017meta, nichol2018first} methods. \textit{Model-based} approaches can store task-specific information in the form of internal representations of tasks via memory components. %As a representative approach of model-based meta-learning,
For example, the memory-augmented neural network (MANN) \cite{santoro2016meta} keeps track of all model training history in external memory and loads the most appropriate model parameters whenever a new task is encountered. 
%For example, CMN \cite{zhu2018compound} employs a key-value memory network schema to yield the optimal video representation in a wider region. 
\textit{Metric-based} models utilize similarity metric functions to compare similarities between new tasks and previously trained tasks with transferable knowledge. For example, the matching network \cite{vinyals2016matching} learns feature spaces across tasks for performing pair-wise comparisons between inputs using an attention mechanism. The recent HTGM \cite{zhang23hierarchical} extends the process of generating task process into a theoretical framework specified by a hierarchy of Gaussian mixture distributions. It learns task embeddings, fits mixed embeddings of tasks, and can score new tasks based on density.
%prototypical network \cite{snell2017prototypical} measures the distance between the support instances' embeddings and their centroids from various classes to perform classification. 
\textit{Optimization-based} models compute gradients from few-shot samples to derive model parameters that can quickly adapt to new tasks. The well-known MAML \cite{finn2017model} optimizes the second-order gradient of meta-tasks to acquire an initialization. Such learned initialization enables the model to converge quickly when fine-tuned with limited labeled examples. %A recent work, MIML \cite{dong2020meta}, further extends MAML by leveraging semantic concepts embedded by BERT as meta information, and achieves promising results on few-shot relation classification. %Note that a recent work MIML \cite{dong2020meta} further extend MAML by introducing meta-information embedded by class name as guide information, and achieve promising results on few-shot relation classification. Meta-SGD \cite{li2017meta} further extends MAML by learning weight initialization, gradient update direction, and learning rate simultaneously in a single step.

\subsection{Few-shot Learning on Graphs} 
Some studies have successfully integrated meta-learning and GNNs to achieve significant improvements in a variety of few-shot situations involving graph-structured data \cite{zhang2022few}, including graph classification \cite{chauhan2020few, yao2020graph, liu2024smart}, link prediction \cite{baek2020learning, zhang2020few}, and node classification \cite{zhou2019meta, liu2020towards, ding2020graph}. %\cite{zhou2019meta, liu2020towards, ding2020graph, huang2020graph, liu2021relative, wen2021metainductive}. Among the models for few-shot node classification, there are three representative ones, namely, Meta-GNN \cite{zhou2019meta}, GPN \cite{ding2020graph} and G-Meta \cite{huang2020graph}.
For example, Meta-GNN \cite{zhou2019meta} leverages MAML to optimize the parameters of the GNN model, GPN enriches prototypical networks with additional graph structure information and G-Meta \cite{huang2020graph} employs local subgraphs to transmit subgraph-specific information for improved model generalization. IA-FSNC \cite{wuinformation} applies data augmentation to few-shot node classification on graphs with parameter fine-tuning, which fine-tunes the initialized parameters of GNNs by utilizing the information from all nodes in the graph to perform support augmentation and query augmentation operations during the meta-learning stage. %Note that other models, such as GFL \cite{yao2020graph} and MI-GNN \cite{wen2021metainductive}, incorporate other prevailing techniques \cite{perez2018film, yoon2019tapnet, vuorio2019multimodal, suo2020tadanet} and achieve better performance. Nevertheless, they focus on multiple networks and are not the same as our scenario. %Furthermore, MI-GNN also employs a similar modulation mechanism to the prior parameters for adapting each graph via conditioning the transformations on input graph representations. The major difference between MI-GNN and our framework in terms of modulation parameters is that we input a per-task embedding rather than a graph embedding.
While the aforementioned models for few-shot node classification have shown promising results, they fail to address the issues (such as the homophilic network assumption and learning stochasticity problems) raised in the introduction.
%In addition, the above models for few-shot node classification cannot address the limitations mentioned in the introduction well, while our framework can alleviate them.

\subsection{Contrastive Learning}
Contrastive learning aims to bring the representations of similar samples closer while pushing away other samples in feature space, enabling the learning of informative representations even in the absence of labeled data. This approach has been demonstrated to be effective in computer vision \cite{chen2020simple, he2020momentum, caron2020unsupervised} and natural language processing \cite{gao2021simcse, liu2024improved, li2024simple}. Additionally, graph contrastive learning has been proven effective in learning discriminative graph embeddings. For example, InfoGCL \cite{xu21infor} propose reducing mutual information between contrasting parts while adhering to the information bottleneck principle at both individual module and overall framework levels, thus minimizing information loss during the graph representation learning process. Notably, contrastive learning is also frequently used to tackle challenges caused by data sparsity. For example, VGCL \cite{yang23generative} and DCCF \cite{ren23disentangle} both employ ingenious methods to generate enhanced views for conducting contrastive learning in recommendation systems, effectively alleviating the issue of data sparsity. While our method aims to leverage this technique to alleviate the issue of data scarcity. Recent studies \cite{khosla2020supervised, gunelsupervised} have extended contrastive learning to the supervised domain by explicitly leveraging label information, achieving remarkable results in various applications. Since contrastive learning and few-shot learning are both learning paradigms designed for label scarcity scenarios, several works \cite{kaomaml, ni2021close} have investigated the relationship between the two and demonstrated that they could mutually benefit each other by incorporating specific mechanisms. %by incorporating specific mechanisms, they can mutually benefit each other. 
Current models incorporating contrastive learning for few-shot node classification tasks mainly focus on using classical graph contrastive learning techniques \cite{tan2022simple, zhou2022task}. These techniques construct positive sample pairs by changing the features or topological information of the original graph to learn high-quality node representations. TRGM \cite{zhou2022task} considers the relationship between tasks by unsupervised contrastive learning to facilitate few-shot node classification, but it requires constructing a task graph for each task, which causes an additional burden of computational complexity. Instead, our model skips the process of constructing contrastive views and performs supervised contrastive learning within the task.%SGL \cite{tan2022simple} 
\section{Problem setup}
\label{sec_3}
To better understand our proposed model, this section first introduces some background knowledge for few-shot node classification. Some important symbols are summarized in Table \ref{symbol}.

\begin{table}[t]
    \centering
    \tiny
    \caption{Descriptions of important symbols}
    \tiny
    \resizebox{0.7\linewidth}{!}{
    \begin{tabular}{c|c}
    \hline
      \textbf{Symbols}   &  \textbf{Description}\\
    \hline
    $\mathcal{G}, \mathcal{V}, \mathcal{E}$     & attributed network, the node set, and the edge set \\
    $\mathbf{X}, \mathbf{A}$                    & feature matrix and adjacency matrix \\
    $\mathcal{Y}_{train}, \mathcal{Y}_{test}$   & seen classes and unseen classes \\
    $\mathcal{T}_{tr}, \mathcal{T}_{te}$        & meta-training and meta-testing tasks \\
    $\mathcal{S}_i, \mathcal{Q}_i$              & support set and query set in task $\mathcal{T}_i$ \\
    $\mathbf{Z}$                                & final node representations  \\
    $\mathbf{P}, \varphi$                       & class prototype vector and class-specific initialized parameter \\
    $\Theta=\{\theta_e, \theta_p\}$             & prior knowledge of Meta-GPS++ \\ 
    $\lambda_i, \mu_i$                          & scaling and shifting vectors of task $\mathcal{T}_i$ \\
    $\Psi=\{\psi_\lambda, \psi_\mu\}$           & parameters of S$^2$ transformation \\
    $\Theta_i$                                  & task-specific parameter for $\mathcal{T}_i$ \\
    $\tilde{\mathbf{Q}}, \text{K}$              & soft label assignment matrix and high-confidence nodes per class \\
    $\alpha, \beta$                             & step size and meta-learning rate \\
    $\xi, \zeta$                                & weight coefficients for contrastive learning and self-training \\
    $\gamma$                                    & regularization coefficient \\
    $\mathcal{L}_{\mathcal{T}_i}^{ce}, \mathcal{L}_{\mathcal{T}_i}^{cl}$           & task loss and contrastive learning loss for task $\mathcal{T}_i$ \\
    $\mathcal{L}_{\mathcal{T}_i}^{st}$          & self-training loss for task $\mathcal{T}_i$ \\
    \hline
    \end{tabular}
    }
    \label{symbol}
\end{table}
Typically, a graph can be denoted as $\mathcal{G}\!=(\mathcal{V}, \mathcal{E}, \mathbf{X}, \mathbf{A})$, where $\mathcal{V}$ represents the set of nodes $\{v_{1}, v_{2}, \dots, v_{n}\}$, while $\mathcal{E}$ is the set of edges $\{e_{1}, e_{2}, \dots, e_{m}\}$. Each node has an initialized feature vector $\mathbf{x}_{i}\! \in \mathbb{R}^{d}$, and the feature matrix of all nodes can be represented as $\mathbf{X}\!=[\mathbf{x}_{1}, \mathbf{x}_{2}, \dots, \mathbf{x}_{n}]^{\mathrm{T}} \!\in\! \mathbb{R}^{n \times d}$. Moreover, the adjacency matrix of a graph can be denoted as $\mathbf{A}\! \in\! \{0, 1\}^{n \times n}$, where $\mathbf{A}_{ij}\!=\!1$ means that there is an edge connecting node $i$ and $j$, otherwise $\mathbf{A}_{ij}\!=0$.

To maintain consistency with previous work \cite{ding2020graph, wang2020graph}, we divide the label set $\mathcal{Y}\!=\{y_{1}, y_{2}, \dots, y_{l}\}$ corresponding to the nodes in the graph $\mathcal{G}$ into two disjoint sets: the \textit{seen} classes $\mathcal{Y}_{train}$ and the \textit{unseen} classes $\mathcal{Y}_{test}$ ($\mathcal{Y}_{train} \cap \mathcal{Y}_{test}\!=\emptyset$). Specifically, there are sufficient labeled nodes %is sufficient information contained in these labeled nodes available for each class 
in $\mathcal{Y}_{train}$, but only limited labeled nodes (\textit{i.e.}, support set $\mathcal{S}$) are available for each class in $\mathcal{Y}_{test}$ \cite{ding2020graph, wang2022task}. Few-shot node classification problems aim to build a classifier that utilizes meta-knowledge gained from the training set $\mathcal{Y}_{train}$ during the meta-training phase to predict the labels of unlabeled nodes (\textit{i.e.}, query set $\mathcal{Q}$) from \textit{unseen} classes in $\mathcal{Y}_{test}$. %that can predict the categories of unlabeled nodes (\textit{i.e.}, query set $\mathcal{Q}$) from \textit{unseen} node classes in $\mathcal{Y}_{test}$ by transferring the meta-knowledge obtained from $\mathcal{Y}_{train}$ during meta-training stage. 
Notably, the classic $N$-way $K$-shot task is defined as randomly selecting $N$ categories from $\mathcal{Y}_{test}$, with each category having $K$ labeled nodes. 

The training of our model adheres to the meta-learning approach, which consists of two stages: \textit{meta-training} and \textit{meta-testing}. To accomplish this, Meta-GPS++ utilizes an episodic training technique that is comparable to those of other meta-learning models \cite{finn2017model}. The main idea behind episodic training is to sample nodes from the training set $\mathcal{Y}_{train}$ to simulate real-world testing scenarios. This appraoch enables the model to learn generalizable knowledge by training across multiple episodes, each comprising numerous tasks. %across many episodes containing numerous tasks. 
The entire meta-training procedure is constructed upon $\Pi$ meta-training tasks, denoted as $\mathcal{T}_{tr}$, each consisting of a support set $\mathcal{S}_i$ and a query set $\mathcal{Q}_i$, constructed for each episode. Specifically, $\mathcal{T}_i$ represents a meta-training task. The meta-training task set $\mathcal{T}_{tr}$ are characterized by $N$-way $K$-shot classification:
\begin{comment}
\begin{equation}
\centering
\begin{aligned}
    \mathcal{T}_{tr} &=\{\mathcal{T}_{i}\}_{i=1}^{\Pi}, \quad \quad \quad \quad \quad \mathcal{T}_i =\{\mathcal{S}_i, \mathcal{Q}_i\}, \\
    \mathcal{S}_i &=\{(v_{i,k}, y_{i,k})\}_{k=1}^{N \times K}, \quad
    \mathcal{Q}_i =\{(\tilde v_{i,k}, \tilde y_{i,k})\}_{k=1}^{N \times M}
\end{aligned}
\end{equation}
\end{comment}

\begin{equation}
\centering
\begin{aligned}
    \mathcal{T}_{tr} &=\{\mathcal{T}_{i}\}_{i=1}^{\Pi}, \quad \quad \mathcal{T}_i =\{\mathcal{S}_i, \mathcal{Q}_i\}, \\
    \mathcal{S}_i &=\{(v_{i,1},y_{i,1}), (v_{i,2}, y_{i,2}), \dots, (v_{i,k}, y_{i,k})\}_{k=1}^{N \times K}, \\
    \mathcal{Q}_i &=\{(\tilde v_{i,1}, \tilde y_{i,1}), (\tilde v_{i,2}, \tilde y_{i,2}), \dots, (\tilde v_{i,k}, \tilde y_{i,k})\}_{k=1}^{N \times M}
\end{aligned}
\end{equation}
where the nodes in the support set $\mathcal{S}_i$ and the query set $\mathcal{Q}_i$ are both sampled exclusively from the training set $\mathcal{Y}_{train}$. Each class within the support set $\mathcal{S}_i$ contains $K$ nodes, while the corresponding class within the query set $\mathcal{Q}_i$ samples $M$ nodes from the remaining data in each class.

The support set $\mathcal{S}_i$ of each meta-task $\mathcal{T}_i$ is used for model training to minimize the prediction loss on the query set $\mathcal{Q}_i$. The model can acquire a large amount of general knowledge during the meta-training phase.
%and will be evaluated for performance on the meta-testing task $\mathcal{T}_{te}=\{\mathcal{S}, \mathcal{Q}\}$, where the composition of $\mathcal{T}_{te}$ follows the same paradigm as $\mathcal{T}_{tr}$, and the only difference is that $\mathcal{T}_{te}$ samples the data with unseen classes from $\mathcal{Y}_{test}$ instead of $\mathcal{Y}_{train}$.
During the meta-testing stage, a meta-testing task $\mathcal{T}_{te}\!=\{\mathcal{S}, \mathcal{Q}\}$ samples data from unseen classes $\mathcal{Y}_{test}$. The support set $\mathcal{S}$ and query set $\mathcal{Q}$ are constructed using the $N$-way $K$-shot method, consistent with the composition of tasks in $\mathcal{T}_{tr}$.%Finally, in the meta-testing phase, a meta-testing task $\mathcal{T}_{te}=\{\mathcal{S}, \mathcal{Q}\}$ samples the data with unseen classes from $\mathcal{Y}_{test}$ instead of $\mathcal{Y}_{train}$, where the composition of support set $\mathcal{S}$ and query set $\mathcal{Q}$ follow the $N$-way $K$-shot manner as those in $\mathcal{T}_{tr}$. Notably, the support set $\mathcal{S}$ with labeled nodes in $\mathcal{T}_{te}$ is used to fine-tune the model for adapting to the $N$ new classes, while the query set $\mathcal{Q}$ with unlabeled nodes is utilized to evaluate the model's performance. %will be evaluated for performance in the query set $\mathcal{Q}$ of the meta-testing task $\mathcal{T}_{te}=\{\mathcal{S}, \mathcal{Q}\}$, where the composition of support set $\mathcal{S}$ and query set $\mathcal{Q}$ both follow the $N$-way $K$-shot manner as $\mathcal{T}_{tr}$, and the only difference is that $\mathcal{T}_{te}$ samples the data with unseen classes from $\mathcal{Y}_{test}$ instead of $\mathcal{Y}_{train}$. the support set $\mathcal{S}$ in $\mathcal{T}_{te}$ is used to fine-tune the model
\section{proposed framework}
\begin{figure*}
    \centering
    \includegraphics[width=0.9\textwidth]{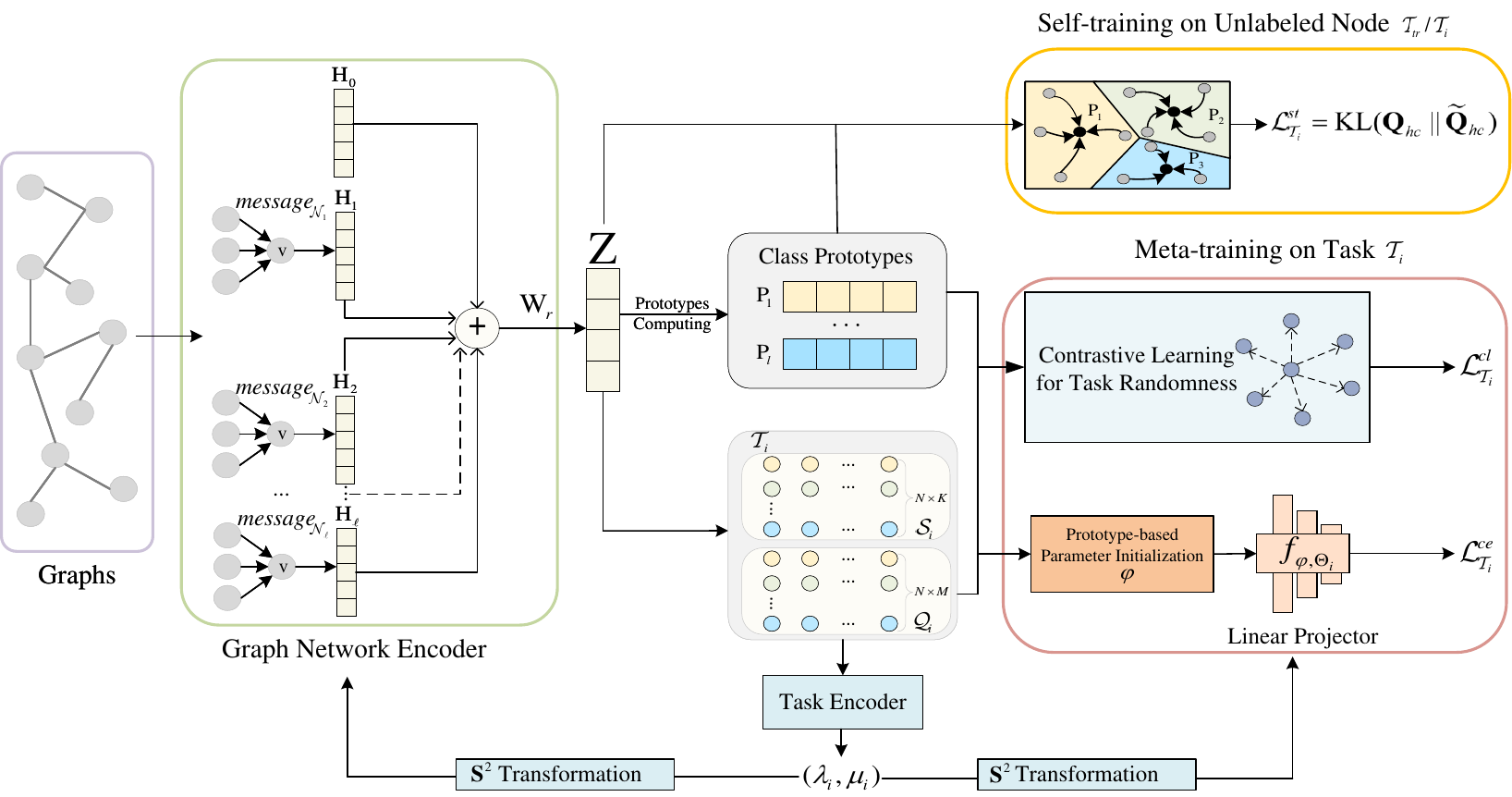}
    \caption{Overall architecture of our framework (best viewed in color). Our model first uses the designed graph layer to encode the given graph to obtain the final node embedding $\mathbf{Z}$ and calculates the corresponding class prototypes. Then, we sample nodes from the graph to form task $\mathcal{T}_i$. These nodes and class prototypes are input into the prototype-based initialization module and contrastive learning for task randomness module. In addition, we feed $\mathcal{T}_i$ into the task encoder to obtain corresponding scaling and translation vectors, which are applied to the prior parameters of the meta-model.}
    \label{framwork}
\end{figure*}

In this section, we provide a detailed introduction to the Meta-GPS++ framework proposed for few-shot node classification on graphs. To better present our model, the overall architecture of Meta-GPS++ is shown in Fig. \ref{framwork}. Subsequently, we provide a detailed explanation of the five critical modules of this basic framework: \textit{a graph network encoder}, \textit{prototype-based parameter initialization}, \textit{contrastive learning for task randomness}, \textit{self-training for model regularization}, and \textit{S$^2$ transformation for different tasks}.

\subsection{Graph Network Encoder}
\label{network_encoder}
Basically, in few-shot node classification tasks, it is crucial to learn informative node representations that incorporate both feature and structural information. Currently, most prevalent GNN models adopt the message passing paradigm, which recursively updates node representations by aggregating information from neighboring nodes. Concretely, this paradigm can be formally expressed as:
\begin{equation}
\label{message_passing}
\begin{aligned}
    s_v^l &= \textbf{AGGREGATE}(\{h_u^{l-1}:u \in \mathcal{N}_v\}) \\
    h_v^l &= \textbf{COMBINE}(h_v^{l-1}, s_v^l)
\end{aligned}
\end{equation}
where $s_v^l$ represents messages that aggregate representations at layer $l$ from the target node's neighbors. $\mathcal{N}_v$ symbolizes the set of neighboring nodes of target node $v$ (\textbf{including node $v$}). %and $h_u^{l-1}$ is the representation of node $u$ at layer $l$-1.  
Additionally, $h_v^l$ denotes the representation of the updated node $v$ at layer $l$ obtained through two key functions, namely, AGGREGATE and COMBINE. There exist numerous GNN architectures \cite{kipf2016semi, velivckovic2017graph, wu2019simplifying} that differ in their selection of these two functions.
A well-known example is simplifying graph convolution (SGC), which is defined as:
\begin{equation}
\label{sgc}
    \mathbf{Z} = \mathbf{\tilde{ \hat A}} \mathbf{\tilde{\hat A}}\cdots \mathbf{\tilde {\hat A}}\mathbf{X}\mathbf W^{(0)}\mathbf W^{(1)}\cdots\mathbf W^{(l-1)}=\mathbf{\tilde{\hat A}}^l\mathbf{X}\mathbf{W^*}
\end{equation}
where $\mathbf{\tilde{\hat{A}}}\!=\mathbf{\hat{D}}^{-1/2}\mathbf{\hat{A}}\mathbf{\hat{D}}^{-1/2}$ is a normalized symmetric adjacency matrix, $\mathbf{\hat{A}}\!=\mathbf{A}+\mathbf{I}$ is an adjacency matrix $\mathbf{A}$ with the added self-loop and $\mathbf{\hat{D}}\!=\mathbf{D}+\mathbf{I}$ is a corresponding diagonal degree matrix with self-loops. $\mathbf{W^*}$ denotes the compressed weight matrix.

The aforementioned GNN models perform well in homophilic graphs but struggle in heterophilic ones. One potential explanation is that these models do not account for whether the connected nodes belong to the same class. Instead, they merge the features of the neighbors with those of the target node in a cumulative manner, resulting in undistinguishable node representations and consequent performance degradation \cite{yan2021two}. Therefore, a natural idea to solve this problem is to update the features of nodes separately rather than updating them together with their neighbors, avoiding the ego-embedding of nodes and the excessive similarity between the embeddings of nodes and their neighbors in heterophilic graphs. Specifically, we substitute $s_v^l$ in Eq. \ref{message_passing} with $s_{v,i}^l\!=\textbf{AGGREGATE}(\{h_u^{l-1}:u\in \mathcal{\widetilde{N}}_i(v)\})$, where $\mathcal{\widetilde{N}}_i(v)$ denotes the set of $i$-hop neighbors of node $v$ (\textbf{excluding node $v$}). Furthermore, to prevent the conflation of a node's individual features with those of its neighbors, the \textbf{COMBINE} function employs a concatenation operator instead of an average \cite{kipf2016semi} or weighted average \cite{velivckovic2017graph}. This approach can be more expressive, as non-aggregated representations can evolve independently over multiple rounds of propagation, resulting in embeddings that are readily distinguishable across different classes of nodes.

\begin{figure*}[ht]
    \centering
    \includegraphics[width=0.5\textwidth]{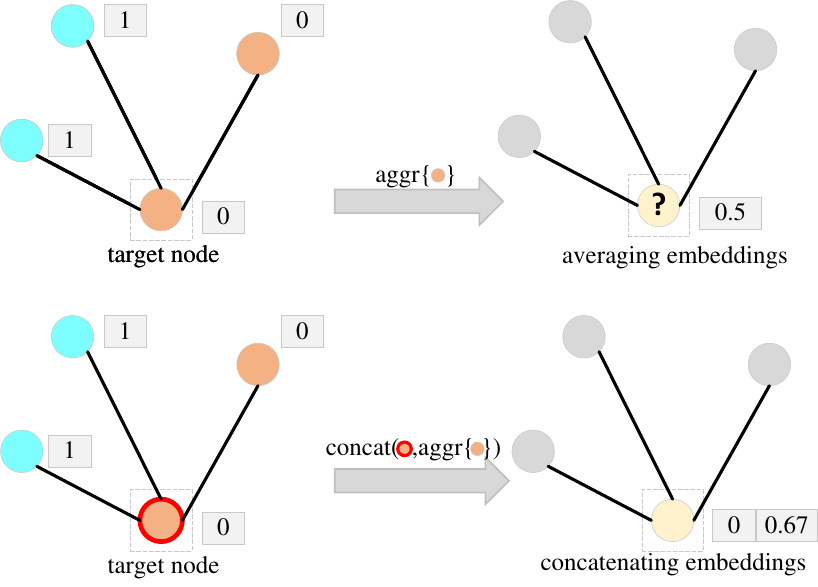}
    \caption{Illustration of averaging or concatenating ego- and neighbor-embeddings. In the upper part, the target node embedding is confusing because ego- and neighbor-embeddings are mixed together. In the lower part, ego-embedding is crucial, and 1-hop embeddings can also help capture the neighboring nodes' label distribution of the target node.}
    \label{concat}
\end{figure*}

To demonstrate the above process more visually, we provide a toy example. From Fig. \ref{concat}, by concatenating the ego-embedding and the aggregated neighbor-embedding for the node of interest instead of averaging them together, we can keep useful information in both the ego-embedding and the aggregated neighbor-embedding. Specifically, since the ego-embedding is not mingled with neighbor-embedding, it keeps the untouched information for itself, which can be decisive; even if our neighborhood aggregator is still the average function, the aggregated neighbor-embedding can keep information about the distribution of the class labels in the 1-hop neighborhood, which can also be helpful for inference of the node of interest.

Drawing inspiration from this, we propose a graph layer that is effective in heterophilic attributed networks and can be mathematically formulated as follows. %regardless of the properties of attributed networks. Specifically, we use a simple but effective method to aggregate features for attributed networks in both a cumulative and non-cumulative manner, and average the features obtained in different manners, respectively. The graph layer can be formalized below:

\begin{equation}
\label{esgc}
  \begin{aligned}
     &\mathbf{F} = \sigma(\mathbf{X}\mathbf{W}_f), \quad
     \mathbf{H}_0 \equiv \mathbf{F}, \quad
     \mathbf{H}_i = \mathbf{\tilde{A}}_i\mathbf{F}, \\
     &\mathbf{R} = {||}_{i=0}^{\ell}\mathbf{H}_i, \quad\, %\in \mathbb{R}^{n \times (l+1) \times d^\prime}
     \mathbf{Z} = \sigma(\mathbf{R}\mathbf{W}_r)
     %\eta = \sigma(\mathbf{R}\mathbf{W}_r), \quad
     %\mathbf{Z} = \text{Squ}(\text{Res}(\eta)\mathbf{R})
  \end{aligned} 
\end{equation}
where $\mathbf{X}\! \in\! \mathbb{R}^{n \times d}$ and $\sigma(\cdot)$ are the initialized node features and the activation function, respectively. $\mathbf{\tilde{A}}_i\!=\mathbf{\bar{D}}_i^{-1/2}\mathbf{\bar{A}}_i\mathbf{\bar{D}}_i^{-1/2}$ denotes the $i$-hop neighbors' normalized symmetric adjacency matrix \textbf{without self-loops}, where $\mathbf{\bar{A}}_i$ represents the adjacency matrix\footnote{For example, $\mathbf{\bar{A}}_1\!=\mathbb{I}(\mathbf{A}-\mathbf{I}>0), \mathbf{\bar{A}}_2\!=\mathbb{I}(\mathbf{A}^2-\mathbf{A}-\mathbf{I}>0)$, where $\mathbb{I}(\cdot)$ is the indicator matrix and $\mathbf{I}$ is the identity matrix.} of $i$-hop neighbors  %In this work, we mainly focus on the neighbors within the second order of the target node.} 
and $\mathbf{\bar{D}}_i$ symbolizes its corresponding diagonal degree matrix. $||$ denotes the concatenation operator. Additionally, $\mathbf{F}, \mathbf{H}_i, \mathbf{Z}\! \in\! \mathbb{R}^{n \times d^{\prime}}$ symbolize the hidden node representations, the aggregated embeddings of $i$-hop neighbors, and the final node embeddings, respectively. %$\mathbf{R}\! \in\! \mathbb{R}^{n \times (\ell+1) \times d^\prime}$ 
$\mathbf{R}$ and $\mathbf{W}_r$ represent the concatenated embeddings and the corresponding weight matrix, respectively. %The attention coefficient $\eta\! \in\! \mathbb{R}^{n \times (\ell+1) \times 1}$ determines how much information from the aggregated representations of different-hop neighbors should be preserved to generate the final representation of each node. %and $\eta \in \mathbb{R}^{n \times (\ell+1) \times 1}$ indicates the attention coefficient to measure how much information of the aggregated neighbors' representations obtained from different-hop neighbors should be maintained to generate the final representation of each node. 
%In addition, Squ and Res are operations that correspond to "squeeze" and "reshape", respectively, to fit the dimensions of the matrix.
%Moreover, Squ and Res represent ``squeeze'' and ``reshape'' operations to match the matrix's dimensions, respectively.

In this approach, we begin with the feature transformation of all nodes without considering structural information. Then, the neighbor-embeddings are aggregated without self-loops by aggregating features of $i$-hop neighbors. Afterward, we concatenate the ego-embeddings and different-hop neighbor-embeddings to acquire the hidden representation $\mathbf{R}$. The rationale behind this is that in heterophilic graphs, nodes and their non-local neighbors are likely to exhibit similar patterns \cite{liu2020non}. Ultimately, we employ an attention mechanism on the hidden representations to extract valuable features that can be utilized for classifying nodes.
%In the above approach, we first perform the feature transformation process for all nodes without considering structure information. Then, the aggregated neighbor-embeddings without ego are derived by aggregating the $i$-hop neighbors' features. These ego-embeddings and different-hop neighbor-embeddings are concatenated to obtain the intermediate representations $\mathbf{R}$. The underlying idea is that nodes and their non-local neighbors are likely to have similar patterns in heterophilic graphs \cite{liu2020non}. Finally, the attention mechanism is applied to the intermediate representation to select informative features for classifying nodes. %As revealed in \cite{wen2021metainductive}, the weight matrix can be applied to similarly aggregate and update messages from neighbors given the feature space. In other words, it can be seen as prior knowledge of the model in meta-learning. 

With the graph encoding mechanism described above, Meta-GPS++ can learn discriminative node embeddings in heterophilic graphs. The prior parameters of the graph network encoder are denoted as $\theta_e=\{\mathbf{W}_f,\mathbf{W}_r\}$.
%By adopting the above approach to encoding networks, our framework can learn more expressive node representations in heterophilic networks. Additionally, we denote the prior parameters of the graph layer as $\theta_e=\{\mathbf{W}_f,\mathbf{W}_r\}$.%Additionally, Meta-GPS can greatly benefit from pre-computing node features and vastly improve efficiency based on the pre-processing step, especially in large graphs.

\subsection{Prototype-based Parameter Initialization}
\label{PI}
The prototype is a representative point that summarizes a group of data belonging to the same category in a more compact form. To derive the prototype vector for each class within $\mathcal{Y}_{train}$, we can compute the mean of the embedded nodes that belong to a particular class, which can be formulated as follows:

\begin{equation}
\label{proto}
    \mathbf{P}_j =  \frac{1}{|\mathcal{V}_j|}\sum\nolimits_{k\in \mathcal{V}_j}\mathbf{Z}_k
\end{equation}
where $\mathbf{Z}_k$ is node $v_k$'s embedding after executing the graph network encoder. $\mathcal{V}_j$ is the set of nodes belonging to class $j$.
%Note that we declare in Section \ref{sec_3} that abundant labeled nodes are available for each class in $\mathcal{Y}_{train}$. Thus, 
The prototype $\mathbf{P}_j$ reflects the unique characteristics of a particular category, which can be seen as the meta-information of that category.

MAML-based graph meta-learning models are deeply affected by stochasticity, which significantly impacts performance. Such models use random and class-independent initialization parameters for all classes. Due to a single shared initialization for tasks encompassing distinct categories of sampled nodes, these models rely heavily on statistics specific to instances and are easily affected by data noise. Owing to the intricate nature of the data distribution in sampling tasks, it is challenging to identify a common parameter initialization to obtain the expected parameters for different tasks. %Thus, different parameters may be required conditioned on various categories, which is tough to find a single initialization near all of the desired parameters. %which is hard to quickly find the optimal parameter for every class via a few gradient steps. 
To achieve better performance, we propose utilizing prototype vectors as auxiliary information to furnish class-specific initialized parameters for guidance. This approach enables our model to efficiently identify the most potential parameters with a few gradient steps. Concretely, we take the prototype $\textbf{P}_j\! \in\! \mathbb{R}^{d^{\prime}}$ of class $j$ as the input of a neural network to obtain class-specific initialized parameters $\varphi_j\! \in\! \mathbb{R}^{d^{\prime}}$ of class $j$, which can be expressed as follows:
\begin{equation}
\label{init}
    \varphi_j=\mathbf{MLP}(\mathbf{P}_j;\theta_p), \text{where} \,\,  j=1,\cdots,N
\end{equation}
where $N$ represents the number of categories in the task (\textit{i.e.}, $N$-way). By default, the neural network we adopt is \textbf{MLP} with parameters $\theta_p$ for the sake of efficiency and simplicity. %Equivalently, other sophisticated networks can be applied, such as convolutional networks or recurrent networks. 
%In this way, the initialized parameters incorporate the prototype information and become category-dependent.
Afterward, the initialized parameters become class-dependent.

$\varphi\! \in\! \mathbb{R}^{N\times d^{\prime}}$ can be directly applied to labeled nodes contained in support set $\mathcal{S}_i$ through a feed-forward network layer, that is, $score\!=\! \text{softmax}(\mathbf{Z}\varphi^{\top}+b)\!\in\! \mathbb{R}^{n \times N}$. The $score$ can be regarded as the probability distribution of these labeled nodes, which can be optimized by gradient descent as follows:
\begin{equation}
\label{phi}
    \varphi_i^{\prime}=\varphi-\alpha\nabla_{\varphi} \mathcal{L}^{ce}_{\mathcal{T}_i}(f(\mathcal{S}_i;\varphi,\Theta))
\end{equation}
where $\Theta\!=\!\{\theta_e, \theta_p\}$ are prior parameters and $\mathcal{L}^{ce}_{\mathcal{T}_i}(\cdot)$ is the cross-entropy loss function mentioned in Eq. \ref{loss} below. Similar to MAML, Eq. \ref{phi} can be easily extended to multiple gradient descents. With the process of prototype-based parameter initialization (PI), the adapted parameters are constrained to closely align with the class prototype, despite the presence of any outliers. %Note that MIML \cite{dong2020meta} leverages semantic concepts embedded by BERT as meta information for relation classification. The difference is that these concepts address textual information located in Euclidean space and introduce external knowledge, while our data lie in non-Euclidean space and Meta-GPS mines its own knowledge only.

\subsection{Contrastive Learning for Task Randomness}
The data for each task are randomly sampled from different categories, leading to a random distribution of nodes. This randomness makes it difficult for the model parameters to adapt to the task $\mathcal{T}_i$, resulting in instability when generating node embeddings. To address this issue, we propose a supervised contrastive learning approach within each task, which acts as a regularization for constraining the distribution of node embeddings per task and improving the overall stability of the base model. Moreover, we add prototype vectors to the positive samples used for contrastive learning to represent the data's global distribution information. This helps to address the problem of the large gap between local and global data distributions within the task. Before performing contrastive learning, we first normalize the obtained node embeddings into unit vectors, %map the obtained node embeddings into the space where the contrastive loss is operated through an MLP with one hidden layer, \textit{i.e.}, $\mathbf{U}\!=\!\mathbf{MLP}(\mathbf{Z})$. 
\textit{i.e.}, $\mathbf{U}\!=\!\mathbf{Z}/||\mathbf{Z}||_2$. Other procedures can be expressed as follows: %Since the data for each task is randomly sampled from different categories, the distribution of nodes we obtain is also random. This increases the difficulty level for model parameters to adapt to task $\mathcal{T}_i$, which in turn leads to instability when generating node embeddings. To this end, we introduce a supervised contrastive learning approach within each task, which serves as a regularization upon the base model for constraining the distribution of node embeddings per task.
\begin{equation}
\label{cl}
    \begin{aligned}
        %\mathbf{U}&=pro(\mathbf{Z}) \\
        \mathcal{L}^{cl}_{i}&=-\frac{1}{|\mathcal{V}_{\mathcal{T}_{i,y}}|}\sum_{+\in ind_+ }\log\frac{\exp{(\mathbf{U}_i\cdot\mathbf{U}_+/\tau)}}{\sum_{k\in ind_k} \exp{(\mathbf{U}_i\cdot\mathbf{U}_k/\tau)}} \\
        \mathcal{L}^{cl}_{\mathcal{T}_i}&=\sum\nolimits_{i\in\mathcal{V}_{\mathcal{T}_i}}\mathcal{L}^{cl}_i
    \end{aligned}
\end{equation}
where $\mathcal{V}_{\mathcal{T}_{i,y}}$ and $\mathcal{V}_{\mathcal{T}_i}$ represent the set of nodes with label $y$ in task $\mathcal{T}_i$, and the set of all nodes in that task, respectively. $ind_+\!=\!\{\mathcal{V}_{\mathcal{T}_{i,y}}\backslash\{i\}\}\cup\{c_i\}$ denotes the indices of nodes in task $\mathcal{T}_i$ that have the same label $y$ as node $i$, but does not include node $i$ itself. $ind_k\!=\{\mathcal{V}_{\mathcal{T}_{i}}\backslash\{i\}\}\cup\{c_i\}$ represents the indices of all nodes in task $\mathcal{T}_i$ except for node $i$. Also, $ind_+$ and $ind_k$ add the index $c_i$ of the corresponding prototype vector. $\tau$ denotes a temperature parameter. %is a temperature hyperparameter that determines the measurement scale of similarity among samples.

Performing supervised contrastive learning can adjust the distribution of sampled data and generate stable node embeddings, which make nodes with the same category gather and nodes with different categories disperse, improving the distinguishability of the model. Moreover, real-world few-shot node classification tasks often involve semantically similar subcategories. For example, in citation networks, \textit{data mining} and \textit{natural language processing} are similar but distinct categories. When nodes from these categories are sampled together in a task, it is difficult to distinguish them. Contrastive learning can explicitly increase the inter-class distance, thus alleviating this issue.

\subsection{Self-training for Model Regularization}
Previous models only consider the nodes selected $\mathcal{V}_{\mathcal{T}_i}$ in each task while directly discarding other nodes in the graph. This may cause the model to overfit to the original classes based on a biased distribution formed by limited samples. Furthermore, graph data are distinct from other data and follow an independent and identically distributed (\textit{i.i.d.}) assumption. Thus, unlabeled nodes $\mathcal{V}_{\mathcal{T}_{tr} /\mathcal{T}_i}$ outside one task can still provide task-relevant and rich self-supervised information, enhancing the meta-learner's adaptability. We propose a self-training approach that fully leverages these unlabeled data. Specifically, we first obtain soft label assignments for unlabeled nodes based on Student's t-distribution \cite{van2008visualizing} with a single degree of freedom, which is defined as:
\begin{equation}
    \tilde{q}_{ij}=\frac{(1+\mid\mid \mathbf{Z}_i-\mathbf{P}_j\mid\mid^2)^{-1}}{\sum_k(1+ \mid\mid\mathbf{Z}_i-\mathbf{P}_k\mid\mid^2)^{-1}}
\end{equation}
where $\tilde{q}_{ij}$ is used to measure the similarity between the embeddings $\mathbf{Z}_i$ of node $i$ and the prototype vector $\mathbf{P}_j$, which can also be interpreted as the probability that node $i$ belongs to category $j$.

Next, we select the top-K predicted values corresponding to nodes in each column of the soft label assignment matrix $\tilde{\mathbf{Q}}\!\in\!\mathbb{R}^{\tilde{n}\times N}$ as high-confidence nodes for that category. %This is because high-confidence nodes are more likely to belong to a certain class, while the remaining low-confidence nodes may contain noise, leading to potential confusion for the model. 
The procedure can be expressed as follows:
\begin{equation}
\label{topk}
    %\mathcal{V}_{hc}=\{v|\text{top-K}(Q_{:,j}), i\in\mathcal{V}_{un}\}
    \tilde{\mathbf{Q}}_{hc}=\text{top-K}(\tilde{\mathbf{Q}},\text{K})_j=\{i|\tilde{\mathbf{Q}}_{ij}\geq \text{sort}(\tilde{\mathbf{Q}}_{:,j})_{\text{K}-1}\}, \, j \in [1,N]
\end{equation}
where sort$(\tilde{\mathbf{Q}}_{:,j})$ represents the result of sorting the $j$-th column of $\tilde{\mathbf{Q}}$ in descending order. $\tilde{\mathbf{Q}}_{hc}$ denotes the probability matrix of these collected nodes. Moreover, we use $\mathcal{V}_{hc}$ to represent the collected high-confidence nodes containing valuable information.

Next, we square the obtained soft label assignments and sharpen them by frequency normalization to acquire the pseudo-target label distribution, as follows:
\begin{equation}
\label{sharpen}
\begin{aligned}
    q_{ij}=\frac{\tilde{q}_{ij}^2/z_j}{\sum\nolimits_{j^\prime}(\tilde{q}_{ij^\prime}^2/z_{j^\prime})}, z_j=\sum\nolimits_i\tilde{q}_{ij}%\\\mathbf{Q}_{hc}&=\text{FN}(\tilde{\mathbf{Q}}_{hc})
\end{aligned}
\end{equation}
where $z_j$ is the soft label frequency. We apply Eq. \ref{sharpen} to $\tilde{\mathbf{Q}}_{hc}$ to obtain the sharpened probability matrix $\mathbf{Q}_{hc}$.
%$\mathbf{Q}_{hc}$ denotes the obtained sharpened probability matrix. 
By doing so, we can boost the high-confidence predictions while reducing low-confidence predictions. %by Eq. \ref{sharpen}. 

Finally, the KL-divergence is employed as a regularization term to optimize the meta-learner by gradually adjusting the probability distribution of $\tilde{\mathbf{Q}}$ for nodes in $\mathcal{V}_{hc}$ to approach the distribution of Q. The specific loss function is defined as:
\begin{equation}
\label{st}
    \mathcal{L}^{st}_{\mathcal{T}_i}=\text{KL}(\mathbf{Q}_{hc}||\tilde{\mathbf{Q}}_{hc})=\sum_i\sum_jq_{ij}\log\frac{q_{ij}}{\tilde{q}_{ij}} 
\end{equation}
%To optimize the meta-learner, we employ the KL-divergence loss function as a regularization term. This gradually adjusts the probability distribution of $\tilde{\mathbf{Q}}$ for nodes in $\mathcal{V}_{hc}$ to approach the Q distribution, thereby promoting more effective learning.

The high-confidence nodes are utilized to regularize the model training, thus reducing the model's dependence on the biased distribution of the few labeled nodes in the task, enabling it to better distinguish nodes from different classes.
\subsection{S$^2$ Transformation for Different Tasks}
Due to significant differences in node characteristics and category distribution among different tasks, directly using $score=\text{softmax} (\mathbf{Z}\varphi^{\top}+b)$ as the final classification output may ignore the differences between tasks. To tackle this issue, we introduce a modulation mechanism that allows for the customization of specific parameters based on different tasks. %to customize specific parameters for different tasks.

%For example, social networks reveal diverse structural patterns for different types of users (\textit{e.g.}, celebrities and regular users) \cite{wen2021metainductive}. Therefore, customizing learnable parameters for each sampled task can effectively capture inter-task differences, which is very beneficial to maximize the model's performance.

%Some researchers have adopted a conditioning mechanism or feature-wise linear modulation (FiLM) \cite{perez2018film} to selectively activate neural networks for the conditioning effects of data from different tasks and have achieved promising results on various applications \cite{yoon2019tapnet, vuorio2019multimodal, suo2020tadanet}. Inspired by FiLM, 
We utilize the S$^2$ transformation to derive a set of parameters that can adapt the previously learned meta-parameters to the requirements of each specific task. To achieve this, we first compute the average embedding of the nodes in the task $\mathcal{T}_i$, which we use to obtain a task-level representation vector $t_i$. This approach brings tasks with similar categories closer in the task representation space. Next, we employ two distinct neural networks to generate the scaling vector $\lambda_i$ and the shifting vector $\mu_i$ to encode the features of the given task. The mathematical formula for this process is as follows:
%We leverage S$^2$ transformation to produce a set of parameters to modulate the prior learned meta-parameters conditioned on a given task. Specifically, we first obtain the task-level representation vector $t_i$ using the mean of the node embeddings contained in a given task $\mathcal{T}_i$, which is simple but effective compared to other expensive methods of obtaining task representations, such as RNNs, CNNs or attention mechanisms. Moreover, in this way, tasks with more similar categories tend to be closer in the task representation space. Then, we implement two separate neural networks to generate the scaling vector $\lambda_i$ and shifting vector $\mu_i$ that encode the characteristics of the given task. The specific formula is given by:
\begin{equation}
\label{s2}
\begin{aligned}
        t_i = \frac{1}{|\mathcal{V}_{t_i}|}\sum\nolimits_{k\in \mathcal{V}_{t_i}}\mathbf{Z}_k, \quad
    \lambda_i = g(t_i;\psi_\lambda), \quad
    \mu_i = g(t_i;\psi_\mu)
\end{aligned}
\end{equation}
where %$\mathcal{V}_{t_i}$ represents the node set, which includes the nodes in $\mathcal{T}_i$. 
$\lambda_i, \mu_i \in \mathbb{R}^{|\Theta|}$ are obtained scaling and shifting vectors, and $|\Theta|$ is the number of parameters in $\Theta$. Moreover, the neural network $g$ is implemented by using two MLPs parameterized by $\psi_\lambda$ and $\psi_\mu$.  %Theoretically, the two separated neural networks $g$ can be arbitrarily parameterized functions. Here, we employ two \textbf{MLPs} in line with the neural network used in \textit{prototype-based parameter initialization}, parameterized by $\psi_\lambda$ and $\psi_\mu$.

Next, we leverage the derived scaling and shifting for modulating the meta-parameters $\Theta$ of the task, defined as:
\begin{equation}
\label{adaption}
    \Theta_i = \lambda_i \odot \Theta + \mu_i
\end{equation}
where $\odot$ denotes element-wise multiplication, as all variables (\textit{i.e.}, $\lambda_i$, $\mu_i$, and $\Theta$) have the same dimension. This implies that each element in $\lambda_i$ and $\mu_i$ controls the weights of neurons specific to each task. By applying this transformation, transferable knowledge is adjusted to the task-specific parameter (\textit{i.e.}, $\Theta_i$), allowing for similar tasks to produce comparable scaling and shifting vectors. Consequently, more information can be shared between similar tasks with ease. 
%In essence, the S$^2$ module does not directly specify the transformation, but it encodes the rules of how to perform the transformation for each task. This is equivalent to a type of hypernetwork \cite{ha2016hypernetworks}, in which the task's prior $\theta$ is adjusted by a secondary network (parameterized by $\Psi$) to accommodate the changing input task. 
Note that during the S$^2$ transformation, only the meta-learned prior parameters $\Theta$ of the task network are updated. The prototype-based parameters $\varphi$ remain fixed. %Note that the prototype-based parameters $\varphi$ are kept fixed in the S$^2$ transformation and that only the meta-learned prior parameters $\Theta$ of the task network are updated.

\subsection{Meta-optimization}
To enhance the adaptability of our model to new tasks with limited labeled data, we employ a meta-learning fashion. Meta-GPS++ is based on MAML and utilizes gradient-based optimization for learning. In this process, we calibrate the prior parameters for each task $\mathcal{T}_i$ by executing several gradient descent steps on the support set $\mathcal{S}_i$. Subsequently, we optimize the prior parameters by backpropagating the task loss computed on the query set $\mathcal{Q}_i$.

\textbf{Meta-training.} To adapt to task $\mathcal{T}_i$, we need to input the nodes in support set $\mathcal{S}_i$ into the model and calculate the task loss $\mathcal{L}^{ce}_{\mathcal{T}_i}$, the contrastive learning loss $\mathcal{L}^{cl}_{\mathcal{T}_i}$, and the self-training loss $\mathcal{L}^{st}_{\mathcal{T}_i}$. Specifically, the cross-entropy loss function used can be depicted by the subsequent formula:
\begin{equation}
\label{loss}
\begin{aligned}
    \mathcal{L}^{ce}_{\mathcal{T}_i}(\mathcal{S}_i,\varphi_i^{\prime},\Theta_i) &= -\sum_{(v_i,y_i)\in \mathcal{S}_i}\Big(y_i\log f(v_i;\varphi^{\prime}_i,\Theta_i) + &(1-y_i)\log(1-f(v_i;\varphi^{\prime}_i,\Theta_i))\Big)    
\end{aligned}
\end{equation}

\begin{algorithm}[h]
\caption{Meta-GPS++ model training}
\label{pseudo}
\KwData{Task Distribution $p(\mathcal{T})$, $\mathcal{G=(\mathcal{V}, \mathcal{E}, \mathbf{X}, \mathbf{A})}$, step sizes $\alpha$, and meta-learning rate $\beta$}
\KwResult{Trained graph meta-learning model Meta-GPS++}

%obtain expressive node representations $\leftarrow$ Sec.\ref{network_encoder}\;
randomly initialize: $\Theta = \{\theta_e, \theta_p\}, \Psi = \{\psi_{\lambda}, \psi_{\mu}\}$\;

\While{not convergence}{
sample batch of tasks $\mathcal{T}_i\sim p(\mathcal{T})$\;
compute each prototype $\mathbf{P}_j$ of $\mathcal{Y}_{train}$ with Eq. \ref{proto}\;
    \For{each task $\mathcal{T}_i$}{
        sample support set $\mathcal{S}_i$ and query set $\mathcal{Q}_i$ for task $\mathcal{T}_i$\;
        obtain node representations of $\mathcal{S}_i$ and $\mathcal{Q}_i$ in Sec.\ref{network_encoder}\;
        %compute each prototype $\mathbf{P}_j$ in $\mathcal{T}_i$\ with Eq. \ref{proto}\;
        \For{j=1 to $N$}{
            $\mathbf{P}_j$-specific parameter initialize $\varphi_{j}$ with Eq. \ref{init}\;
        }
        compute updated $\varphi$ %with gradient descent 
        with Eq. \ref{phi}\; %via $\mathcal{S}_i$
        %$\varphi_i^{\prime}=\varphi-\alpha\nabla_{\varphi} \mathcal{L}_{\mathcal{T}_i}(f(\mathcal{S}_i;\varphi,\Theta))$\;
        compute contrastive learning loss with Eq. \ref{cl}\;
        compute self-training loss with Eq. \ref{st}\;
        compute S$^2$ vectors $\lambda_i, \mu_i$ with Eq. \ref{s2}\;
        obtain $\Theta_i$ to suit $\mathcal{T}_i$ with Eq. \ref{adaption}\;
        compute support loss using $\mathcal{S}_i$ with Eq. \ref{loss}\; %calculate $\nabla_{\Theta_i}\mathcal{L}_{\mathcal{T}_i}(f(\mathcal{S}_i;\varphi_i^{\prime},\Theta_i))$ using $\mathcal{S}_i$ with Eq. \ref{loss}\;
        compute adapted parameter $\Theta_i^{\prime}$ with Eq. \ref{update}\;
        compute query loss using $\mathcal{Q}_i$\;
    }
    update $\Theta, \Psi$ by backpropagation of total query loss with Eq. \ref{meta-update}.%with total query loss with Eq. \ref{meta-update}.
    %$\Theta=\Theta-\beta\nabla_{\Theta}\mathcal{L}(f_{\varphi^{\prime},\Theta}, g_\Psi)$,$\Psi=\Psi-\beta\nabla_{\Psi}\mathcal{L}(f_{\varphi^{\prime},\Theta}, g_\Psi)$
}
return the trained Meta-GPS++.
\end{algorithm}

Here, $\varphi_i^{\prime}$ represents the optimized prototype-based initialized parameters as mentioned in Eq. \ref{phi}, while $\Theta_i$ denotes the parameters tailored to task $\mathcal{T}_i$. Furthermore, the output layer of Meta-GPS++, which is parameterized by $\varphi^{\prime}_i$ and $\Theta_i$, is denoted as $f(\ast;\varphi^{\prime}_i,\Theta_i)$.

To tailor the parameters $\Theta_i$ for a specific task $\mathcal{T}_i$, we update them to $\Theta_i^{\prime}$ using a single gradient descent step, guided by the support task loss mentioned earlier. %We suit $\Theta_i$ to task $\mathcal{T}_i$ by updating it to $\Theta_i^{\prime}$ using one gradient descent step, based on the support task loss mentioned above. 
The process is as follows:

\begin{equation}
\label{update}
    \Theta_i^{\prime} = \Theta_i - \alpha\nabla_{\Theta_i}\mathcal{L}^{ce}_{\mathcal{T}_i}(f(\mathcal{S}_i;\varphi_i^{\prime},\Theta_i))
\end{equation}
where $\alpha$ is the meta-step size. Notably, we can easily extend Eq. \ref{update} to multi-step gradient updates.

Once we obtain the updated $\Theta_i^{\prime}$, we can use it to calculate the loss for query set $\mathcal{Q}_i$ included in per task $\mathcal{T}_i$, \textit{i.e.}, $\mathcal{L}^{ce}_{\mathcal{T}_i}$ ($\mathcal{Q}_i,\varphi_i^{\prime},\Theta_i$). Our ultimate goal is to minimize the meta-objective function under the distribution of meta-training tasks, defined as:
\begin{equation}
\label{objective}
\begin{aligned}
    \min\limits_{\Theta,\Psi}\mathcal{L}(f_{\varphi^{\prime},\Theta}, g_\Psi) = \min\limits_{\Theta,\Psi}\sum_{\mathcal{T}_i\sim p(\mathcal{T})} [ \mathcal{L}^{ce}_{\mathcal{T}_i}(f(\mathcal{Q}_i;\varphi_i^{\prime},\Theta_i^\prime)) + \xi \mathcal{L}^{cl}_{\mathcal{T}_i} + \zeta \mathcal{L}^{st}_{\mathcal{T}_i}] + \gamma \left\|\Psi\right\|_2^2
\end{aligned}
\end{equation}
where $\Psi=\{\psi_\lambda, \psi_\mu\}$ and $\xi, \zeta, \gamma$ are weight coefficients used to control the impact of parameter regularization.

We perform meta-optimization across tasks using stochastic gradient descent and update the model parameters accordingly:
\begin{equation}
\label{meta-update}
    \Theta=\Theta-\beta\nabla_{\Theta}\mathcal{L}(f_{\varphi^{\prime},\Theta}, g_\Psi), \quad
    \Psi=\Psi-\beta\nabla_{\Psi}\mathcal{L}(f_{\varphi^{\prime},\Theta}, g_\Psi)
\end{equation}
where $\beta$ symbolizes the meta-learning rate. The specific training process of the proposed Meta-GPS++ can be found in Algorithm \ref{pseudo}.

\textbf{Meta-testing.}
In the meta-testing phase, we use the same procedure as in meta-training. This involves adapting the learned prior parameters $\Theta$ and $\Psi$ to the support set $\mathcal{S}$ for the given meta-testing task $\mathcal{T}_{te}$. We achieve this by performing one or a few gradient descent steps. However, this process does not include contrastive learning and self-training. Subsequently, we determine the final model performance by evaluating its performance on the query set $\mathcal{Q}$ in task $\mathcal{T}_{te}$. %During the meta-testing phase, we generally follow the same procedure as during the meta-training. Specifically, the learned prior parameters $\Theta, \Psi$ are adapted to support set $\mathcal{S}$ in meta-testing task $\mathcal{T}_{te}$ via one or a small number of gradient descent steps. Then, we evaluate the model's performance on the query set $\mathcal{Q}$ %The only difference from meta-training is that there are only a few labeled nodes in each class, such that we use their average representations as the class prototype. Subsequently, the learned prior parameters $\Theta, \Psi$ are further adapted to support set $\mathcal{S}$ in meta-testing task $\mathcal{T}_{te}$ and evaluate the model's performance on the query set $\mathcal{Q}$ %contained unlabeled nodes in $\mathcal{T}_{te}$.

\subsection{Complexity Analysis}
%\textcolor{blue}{Our model primarily comprises five components, and we will proceed to analyze the time complexity required for each component in detail. In the graph network encoder, the time complexity is $O(ndd^\prime+\ell n^2d^\prime+n(\ell+1)d^{\prime2})$, where $n$ and $\ell$ denote the number of nodes and graph layers. $d$ and $d^\prime$ denote the dimension of original and hidden features. The time complexity for prototype-based parameter initialization is $O(Nd^{\prime2})$, where $N$ is the number of categories. The time complexity of contrastive learning is $O(NK(NK-1)d^\prime)$, where $K$ is the number of shots. Next, the time complexity of self-training is $O(nN+n^2)$. The S$^2$ transformation of time complexity is $O(nd^{\prime2})$. The overall time complexity is acceptable to us.}

Our model consists of five primary components, each with its detailed time complexity analysis. Beginning with the graph network encoder, its time complexity is represented as $O(ndd^\prime+\ell n^2d^\prime+n(\ell+1)d^{\prime2})$, where $n$ and $\ell$ correspond to the number of nodes and graph layers respectively. Additionally, $d$ and $d^\prime$ signify the dimensions of original and hidden features. Moving on to prototype-based parameter initialization, the time complexity is $O(Nd^{\prime2})$, where $N$ denotes the number of categories. For contrastive learning, the time complexity is $O(NK(NK-1)d^\prime)$, where $K$ represents the number of shots. Subsequently, the time complexity for self-training is $O(nN+n^2)$, while the S$^2$ transformation yields a time complexity of $O(nd^{\prime2})$. Overall, we find the cumulative time complexity within acceptable bounds for our purposes.

\subsection{Discussion}
Since our work is about few-shot learning, there will be different scenarios (\textit{i.e.}, heterogeneous/homogenous and directed/undirected graphs) where different types of edges exist Although in this work we apply the proposed model to homogeneous graphs, this does not mean that it is not applicable in other scenarios. We simply need to modify the graph layers in the network encoder phase to obtain the corresponding node representations, and the model still achieves satisfactory results.

\section{Experiments}
%\textbf{Datasets.} 
\subsection{Dataset}
To extract diverse meta-knowledge, FSL models require a significant number of tasks that involve nodes from different classes during the meta-training phase. Hence, node classification benchmark datasets with only a few classes are not appropriate for this purpose.
\begin{table}[ht]
\centering
%\tiny
\caption{Statistics of the evaluation datasets}
\label{dataset}
%\resizebox{0.9\textwidth}{!}{%
\begin{tabular}{lccccc}
\hline
Datasets            & \# Nodes                     &\# Edges                      &\# Features                 &\# Labels                 &\# Homophily             \\ \hline
Reddit             & 232,965                    & 11,606,919                  & 602                       & 41                      & 0.81                     \\
ogbn-products      & 2,449,029                  & 61,859,140                  & 100                       & 47                      & 0.83                     \\
Amazon-Clothing    & 24,919                     & 91,680                      & 9,034                     & 77                      & 0.62                     \\
Cora-Full          & 19,793                     & 65,311                      & 8,710                     & 70                      & 0.59                     \\
Amazon-Electronics & 42,318                     & 43,556                      & 8,669                     & 167                     & 0.39                     \\
DBLP               & 40,672                     & 288,270                     & 7,202                     & 137                     & 0.29 \\ \hline
\end{tabular}%
%}
\end{table}

%Since FSL models need to be trained with a large number of tasks consisting of nodes from various categories in the meta-training phase to extract effective meta-knowledge, the widely used node classification benchmark datasets containing few categories are not suitable for this work. 
Here, we use six real-world datasets in our experiments for few-shot node classification, which have been used in previous research. These datasets are described in detail as follows.
%In our experiments, we employ six real-world datasets %, namely, \textbf{Reddit} \cite{hamilton2017inductive}, \textbf{ogbn-products} \cite{hu2020open}, \textbf{Amazon-Clothing} \cite{mcauley2015inferring}, \textbf{Cora-Full} \cite{bojchevski2017deep}, \textbf{Amazon-Electronics} \cite{mcauley2015inferring}, and \textbf{DBLP} \cite{tang2008arnetminer}, for few-shot node classification used in earlier research \cite{ding2019deep, wang2020graph, ding2020graph}, which are described in detail as follows:

(I) \textbf{Reddit} \cite{hamilton2017inductive} is a collection of posts from the Reddit forum, forming a post-to-post graph. Each node in the graph represents a post, and these nodes are linked if they have been commented on by the same person. The labels of posts indicate their respective communities. Additionally, 300-dimensional GloVe word vectors that are readily available off-the-shelf define the attributes of each node. We utilize 21/10/10 classes as train/validation/test sets.

%\textbf{Reddit} \cite{hamilton2017inductive} is a post-to-post graph extracted from the Reddit internet forum. Nodes represent posts, and two posts are linked if they are both commented by the same person. Additionally, the label for each node is the community to which the corresponding post belongs. The attributes of nodes are embedded by off-the-shelf 300-dimensional GloVe word vectors. %Following the same splits of previous research \cite{ding2020graph}, we utilize 21/10/10 classes as train/validation/test sets.
(II) \textbf{ogbn-products} \cite{hu2020open} is a co-purchasing network that features products developed by the Open Graph Benchmark (OGB). Each node in the network represents a product, and an edge between two nodes indicates that they are commonly purchased together. The label assigned to each node corresponds to its respective product category. Furthermore, the features of each node are derived from their corresponding product descriptions using the bag-of-words approach. We split the classes as train/validation/test (21/16/10) sets for accommodating the few-shot node classification scenario.
%is a product co-purchasing network from Open Graph Benchmark (OGB). In this network, nodes represent products, and the presence of edges between two products indicates that they are purchased jointly. The node label is the product category. Moreover, the features of nodes are obtained by bag-of-words of the product description. %For this dataset, we remove 6 categories with fewer nodes and split the remaining classes as train/validation/test (21/10/10) sets for accommodating the few-shot node classification scenario.

(III) \textbf{Amazon-Clothing} \cite{mcauley2015inferring} is a network of products that includes items from the Amazon categories of Clothing, Shoes, and Jewelry. Each product is represented as a node in the network, with its features derived from its description. Two nodes are connected if they have been viewed by the same user. Furthermore, each node is labeled based on its fine-grained product subcategory. We use 40/17/20 classes as train/validation/test sets.
% is an Amazon product network made up of items from the categories of ``Clothing, Shoes and Jewelry''. Each product is regarded as a node, and the node features are built using its description. The edges between two products denote that they are viewed by the same user. In addition, the fine-grained product categories determine the node labels. %We use 40/17/20 classes as train/validation/test sets, which is the same as \cite{ding2020graph}.

(IV) \textbf{Cora-Full} \cite{bojchevski2017deep} is the complete version of the well-known Cora dataset, forming a paper citation network. In this dataset, each node corresponds to a paper and every edge represents a citation between nodes. The attributes of each node are created using the bag-of-words technique for both the title and abstract of the corresponding paper, and the labeling of nodes is based on their respective topics. Furthermore, we use 40/15/15 node classes as train/validation/test sets for adapting the scenario of few-shot learning.
%a paper citation network, a full version of well-known Cora dataset. In this dataset, nodes represent papers and edges represent citation links. The node is labeled based on the paper topic. Node attributes are generated by using bag-of-words for the title and abstract of the paper. %Furthermore, we use 40/15/15 node classes as train/validation/test sets for adapting the scenario of few-shot learning.

(V) \textbf{Amazon-Electronics} \cite{mcauley2015inferring} is a subset of Amazon's Electronics products. It is similar to Amazon-Clothing: nodes represent individual products and edges connect two products that are purchased jointly. Node features are generated from the product descriptions, and node labels represent the corresponding low-level product classes. We use 90/37/40 node classes for training/validation/test.
% is a network within the "Electronics" category of Amazon products. Similar to `Amazon-Clothing', nodes mean products, and edges between two products are created if they are bought together. The product description is used to generate node features. Node class labels are made up of low-level product categories. %Following the settings of this dataset in \cite{ding2020graph}, we use 90/37/40 node classes for training/validation/test.

(VI) \textbf{DBLP} \cite{tang2008arnetminer} is a citation network that treats papers as nodes and their citation links as edges. Node features are generated based on the abstracts of papers by leveraging the bag-of-words approach, and the labels represent the venues of papers. Its train/validation/test split is 80/27/30.

% is also a citation network where papers are considered nodes and the link is a citation relationship of papers. The abstracts of papers are utilized to create node features, and the papers are labeled with the venues. %We use this dataset that has been preprocessed by \cite{ding2020graph} and is consistent with its train/validation/test split (\textit{i.e.}, 80/27/30).

The same train/validation/test splits as in previous studies \cite{ding2020graph, wang2020graph} are utilized for the above datasets. Table \ref{dataset} contains the detailed statistics, where ``\#Homophily'' represents the homophily level in a network, measured by the average of node homophily $\mathbf{H}=\frac{1}{|\mathcal{V}|}\sum\nolimits_{v\in\mathcal{V}}\frac{|\{(u,v):u\in\mathcal{N}_v\wedge y_u=y_v\}|}{|\mathcal{N}_v|}$, defined in \cite{pei2020geom}. 
% For these datasets, we use the same train/validation/test splits as in previous studies \cite{ding2020graph, wang2020graph}. The comprehensive statistics of the above datasets can be found in Table \ref{dataset}. Note that ``\#Homophily'' in Table \ref{dataset} is used to measure the homophily level in a network. Here, we adopt the node homophily $\mathbf{H}=\frac{1}{|\mathcal{V}|}\sum\nolimits_{v\in\mathcal{V}}\frac{|\{(u,v):u\in\mathcal{N}_v\wedge y_u=y_v\}|}{|\mathcal{N}_v|}$, defined in \cite{pei2020geom}, where higher values denote stronger homophily and lower values indicate stronger heterophily.

%\textbf{Baselines.} 
\subsection{Baselines}
To demonstrate the effectiveness of our proposed model, we select a series of baseline models for comparison. The following is a detailed introduction to these models. %To verify the superiority of our proposed model, a series of baseline models are selected for comparison. Detailed information on the baseline models compared is listed below.
 
 \noindent $\bullet$ \textbf{DeepWalk} \cite{perozzi2014deepwalk}: It first uses the random walk algorithm to extract node sequences from the graph and then applies the skip-gram model to learn meaningful node embeddings, which explicitly encode the social relations in the continuous real-valued embedding space.
 
 \noindent $\bullet$ \textbf{node2vec} \cite{grover2016node2vec}: It is an improved version of the DeepWalk model that incorporates biased random walks, which are designed to capture both depth-first and breadth-first sampling characteristics, resulting in more effective network embeddings.
 
 \noindent $\bullet$ \textbf{GCN} \cite{kipf2016semi}: It defines the graph convolution operation in the spectral domain by utilizing the first-order approximation of the Chebyshev polynomial to learn low-dimensional node embeddings, and shows competitive performance on node classification tasks. %It leverages the first-order approximation of the Chebyshev polynomial to mimic the graph convolution for learning latent node embeddings.  %and shows competitive performance on node classification task.
 
 \noindent $\bullet$ \textbf{SGC} \cite{wu2019simplifying}: It is the simplified version of GCNs by removing the layer-wise non-linear activation and converting the convolution function into a linear transformation, while still preserving the ability to learn effective node embeddings for the downstream tasks.
 
 \noindent $\bullet$ \textbf{Protonet} \cite{snell2017prototypical}: It is a metric-based meta-learning method that maps the selected samples to a feature space and calculates the average value of each category as the prototype for the corresponding category, then assigns the test samples to the closest prototype.%Prototypical network is a classical metric-based meta-learning method. It maps the sample data of each class into a space and computes their mean to represent the class as a prototype.
 
 \noindent $\bullet$ \textbf{MAML} \cite{finn2017model}: It is an optimization-based meta-learning method that can quickly adapt to new tasks with only limited labeled data by performing one or more gradient descent steps on the random initial parameters of the model.
 
 \noindent $\bullet$ \textbf{Meta-GNN} \cite{zhou2019meta}: It applies the transferable knowledge learned from many similar meta-tasks to unseen tasks with only limited labeled nodes by directly combining GNNs with MAML, obtaining promising performance in few-shot node classification tasks.
 
 \noindent $\bullet$ \textbf{GPN} \cite{ding2020graph}: It uses a GNN-based encoder and evaluator to determine the importance of each node and learn the prototype for each class. To predict labels, GPN calculates the Euclidean distance between a node and its corresponding prototype.
 
 \noindent $\bullet$ \textbf{G-Meta} \cite{huang2020graph}: The fundamental idea of G-Meta is to build a subgraph for each node and utilize local subgraphs to transfer subgraph-specific information and learn general knowledge quickly by means of meta gradients, and the predictions are concentrated in the subgraph surrounding the target node.

 \noindent $\bullet$ \textbf{TENT} \cite{wang2022task}: It proposes a task-adaptive framework including node-level, class-level and task-level adaptations for alleviating the impact of task variance. Particularly, it learns node and class embeddings via node-level and class-level adaptions, and maximizes the mutual information between support set and query set via task-level adaptions.

 \noindent $\bullet$ \textbf{IA-FSNC} \cite{wuinformation}: It introduces data augmentations by leveraging the information of all nodes in the graph, including support augmentation and query augmentation in the parameter fine-tuning stage for good model generalization.  

 \noindent $\bullet$ \textbf{TRGM} \cite{zhou2022task}: It combines graph meta-learning and graph contrastive learning to capture the task-level relations, including task-correlation and task-discrepancy, by introducing the auxiliary constructed different task graphs for few-shot node classification.
 
 We mainly divide the above baseline models into four categories: (I) \textit{Random walk-based models}, including two unsupervised methods: DeepWalk \cite{perozzi2014deepwalk} and node2vec \cite{grover2016node2vec}. After obtaining node representations, we perform logistic regression for node classification. (II) \textit{GNN-based models}, consisting of GCN \cite{kipf2016semi} and SGC \cite{wu2019simplifying}. (III) \textit{Classic meta-learning models}, containing Protonet \cite{snell2017prototypical} and MAML \cite{finn2017model} mainly for \textit{i.i.d} data. (IV) \textit{Graph meta-learning models}, including Meta-GNN \cite{zhou2019meta}, GPN \cite{ding2020graph}, G-Meta \cite{huang2020graph}, TENT \cite{wang2022task}, IA-FSNC \cite{wuinformation}, and TRGM \cite{zhou2022task}, designed specifically for graphs. To adapt to the focused task, we adopt the experimental settings of Meta-GNN in the first two types of baseline models. We also conduct a small grid search for learning rates and hidden feature dimensions within the ranges [1e-3, 2e-3, 1e-2] and [16, 32, 64] respectively. Additionally, we utilize the default parameters in the original papers for the remaining baseline models.

\begin{table*}[ht]
\centering
%\tiny
\caption{Results of the Accuracy and Macro-F1 score on various datasets. Underline: runner-up. $\ast$: out of memory.} %Ab Impro: absolute improvement.}
\resizebox{0.9\textwidth}{!}{%
\begin{tabular}{l|cccccccc|cccccccc}
\hline
\multicolumn{1}{c|}{\multirow{3}{*}{Models}} & \multicolumn{8}{c|}{Reddit}                                                                                                                  & \multicolumn{8}{c}{ogbn-products}                                                                                                            \\ \cline{2-17} 
\multicolumn{1}{c|}{}                        & \multicolumn{2}{c}{5-way 3-shot} & \multicolumn{2}{c}{5-way 5-shot} & \multicolumn{2}{c}{10-way 3-shot} & \multicolumn{2}{c|}{10-way 5-shot} & \multicolumn{2}{c}{5-way 3-shot} & \multicolumn{2}{c}{5-way 5-shot} & \multicolumn{2}{c}{10-way 3-shot} & \multicolumn{2}{c}{10-way 5-shot} \\ \cline{2-17} 
\multicolumn{1}{c|}{}                        & ACC             & F1             & ACC             & F1             & ACC             & F1              & ACC              & F1              & ACC             & F1              & ACC             & F1             & ACC             & F1              & ACC             & F1              \\ \hline
DeepWalk                                     & 0.267           & 0.261          & 0.301           & 0.297          & 0.176           & 0.171           & 0.188            & 0.186           & 0.249           & 0.241           & 0.279           & 0.264          & 0.167           & 0.155           & 0.192           & 0.179           \\
node2vec                                     & 0.271           & 0.256          & 0.312           & 0.298          & 0.198           & 0.186           & 0.234            & 0.226           & 0.253           & 0.247           & 0.305           & 0.299          & 0.176           & 0.163           & 0.212           & 0.202           \\ \hline
GCN                                          & 0.388           & 0.381          & 0.455           & 0.441          & 0.290           & 0.270           & 0.357            & 0.324           & 0.341           & 0.326           & 0.394           & 0.375          & 0.256           & 0.259           & 0.302           & 0.295           \\
SGC                                          & 0.444           & 0.421          & 0.468           & 0.425          & 0.297           & 0.268           & 0.316            & 0.277           & 0.375           & 0.366           & 0.412           & 0.406          & 0.294           & 0.301           & 0.337           & 0.331           \\ \hline
Protonet                                     & 0.346           & 0.333          & 0.376           & 0.364          & 0.198           & 0.180           & 0.233            & 0.214           & 0.291           & 0.286           & 0.305           & 0.297          & 0.216           & 0.195           & 0.276           & 0.263           \\
MAML                                         & 0.291           & 0.268          & 0.311           & 0.291          & 0.152           & 0.122           & 0.179            & 0.156           & 0.351           & 0.343           & 0.402           & 0.391          & 0.262           & 0.241           & 0.294           & 0.270           \\ \hline
Meta-GNN                                     & 0.608           & 0.583          & 0.627           & 0.612          & 0.449           & 0.421           & 0.515            & 0.471           & 0.492           & 0.504           & 0.523           & 0.514          & 0.347           & 0.349           & 0.401           & 0.396           \\
GPN                                          & 0.655           & 0.662          & 0.684           & 0.690           & 0.534           & 0.558           & 0.577            &  0.592           & 0.556           & 0.532           & 0.572           & 0.566          & 0.396           & 0.392           & 0.425           & 0.412           \\
G-Meta                                       & 0.726
& 0.721
& 0.735
& 0.731
& 0.583
& 0.571
& 0.621
& 0.616
& 0.605
& 0.599
& 0.624
& 0.611
& 0.461
& 0.452
& 0.491
& 0.480     \\
TENT                                         & $\ast$           & $\ast$          & $\ast$           & $\ast$          & $\ast$           & $\ast$           & $\ast$            & $\ast$           & $\ast$           & $\ast$           & $\ast$           & $\ast$          & $\ast$           & $\ast$           & $\ast$           & $\ast$           \\ 
IA-FSNC                                         & 0.730           & 0.725          & 0.739           & 0.725          & 0.615           & 0.602           & 0.632            & 0.620           & 0.635           & 0.613           & 0.644  & 0.625         & 0.491          & 0.492           & 0.517                      & 0.510           \\ 
TGRM                                         & 0.739           & 0.728          & 0.751           & 0.732         & \underline{0.633}           & \underline{0.612}           & 0.639            & 0.625           & \underline{0.655}           & 0.630           & \underline{0.671}           & 0.652          & \underline{0.519}           & \underline{0.513}           & \underline{0.533}           & 0.520           \\ \hline
Meta-GPS                                  & \underline {0.766}           & \underline{0.753}          & \underline{0.779}           & \underline{0.772}         & 0.627           & 0.606           & \underline{0.655}            & \underline{0.651}           & 0.654           & \underline{0.646}           & 0.665           & \underline{0.653}          & 0.509           & 0.497           & 0.531           & \underline{0.522} \\
Meta-GPS++                                          & \textbf{0.806}  & \textbf{0.793} & \textbf{0.822}  & \textbf{0.810} & \textbf{0.676}  & \textbf{0.652}  & \textbf{0.692}   & \textbf{0.690}  & \textbf{0.694}  & \textbf{0.682}  & \textbf{0.701}  & \textbf{0.695} & \textbf{0.552}  & \textbf{0.546}  & \textbf{0.561}  & \textbf{0.552}  \\ \hline
%\multicolumn{1}{c|}{Ab Impro}                 & 4\%              & 3.2\%           & 4.4\%            & 4.1\%           & 4.4\%            & 3.5\%            & 3.4\%             & 3.5\%            & 4.9\%            & 4.7\%           & 4.1\%            & 4.2\%           & 3.8\%            & 4.5\%            & 4\%              & 4.2\%            \\ \hline
\end{tabular}%
}

\centering
\resizebox{0.9\textwidth}{!}{%
\begin{tabular}{l|cccccccc|cccccccc}
\hline
\multicolumn{1}{c|}{\multirow{3}{*}{Models}} & \multicolumn{8}{c|}{Amazon-Clothing}                                                                                                         & \multicolumn{8}{c}{Cora-Full}                                                                                                               \\ \cline{2-17} 
\multicolumn{1}{c|}{}                        & \multicolumn{2}{c}{5-way 3-shot} & \multicolumn{2}{c}{5-way 5-shot} & \multicolumn{2}{c}{10-way 3-shot} & \multicolumn{2}{c|}{10-way 5-shot} & \multicolumn{2}{c}{5-way 3-shot} & \multicolumn{2}{c}{5-way 5-shot} & \multicolumn{2}{c}{10-way 3-shot} & \multicolumn{2}{c}{10-way 5-shot} \\ \cline{2-17} 
\multicolumn{1}{c|}{}                        & ACC             & F1             & ACC             & F1             & ACC             & F1              & ACC             & F1              & ACC             & F1              & ACC             & F1             & ACC             & F1              & ACC             & F1              \\ \hline
DeepWalk                                     & 0.367           & 0.363          & 0.465           & 0.466          & 0.213           & 0.191           & 0.353           & 0.329           & 0.236           & 0.221           & 0.259           & 0.247          & 0.153           & 0.132           & 0.170           & 0.157           \\
node2vec                                     & 0.362           & 0.358          & 0.419           & 0.407          & 0.175           & 0.151           & 0.326           & 0.302           & 0.237           & 0.219           & 0.254           & 0.235          & 0.139           & 0.125           & 0.152           & 0.146           \\ \hline
GCN                                          & 0.543           & 0.514          & 0.593           & 0.566          & 0.413           & 0.375           & 0.448           & 0.403           & 0.346           & 0.339           & 0.398           & 0.376          & 0.292           & 0.284           & 0.341           & 0.352           \\
SGC                                          & 0.568           & 0.552          & 0.622           & 0.615          & 0.431           & 0.416           & 0.463           & 0.447           & 0.395           & 0.402           & 0.445           & 0.432          & 0.351           & 0.336           & 0.395           & 0.389           \\ \hline
Protonet                                     & 0.537           & 0.536          & 0.635           & 0.637          & 0.415           & 0.419           & 0.448           & 0.462           & 0.336           & 0.311           & 0.365           & 0.339          & 0.249           & 0.245           & 0.272           & 0.261           \\
MAML                                         & 0.552           & 0.545          & 0.661           & 0.678          & 0.456           & 0.433           & 0.468           & 0.456           & 0.371           & 0.352           & 0.475           & 0.461          & 0.266           & 0.259           & 0.316           & 0.312           \\ \hline
Meta-GNN                                     & 0.721           & 0.726          & 0.748           & 0.775          & 0.614           & 0.597           & 0.642           & 0.629           & 0.522           & 0.510           & 0.591           & 0.571          & 0.472           & 0.461           & 0.533           & 0.526           \\
GPN                                          & 0.724           & 0.732    & 0.761           & 0.790    & 0.650           & 0.661     & 0.677           & 0.679     & 0.532           & 0.523           & 0.603           & 0.594          & 0.509           & 0.491           & 0.562           & 0.556           \\
G-Meta                                       & 0.732     & 0.731          & 0.766     & 0.786          & 0.663     & 0.657           & 0.680     & 0.671           & 0.575     & 0.569     & 0.624     & 0.616    & 0.539     & 0.526     & 0.581     & 0.575     \\
TENT                                         & \underline{0.814}           & 0.795          & 0.822           & 0.811          & \underline{0.735}           & 0.712           & \underline{0.745}            & 0.732           & 0.648           & 0.633            & 0.692           & 0.681          & 0.517           & 0.504           & 0.560           & 0.557           \\ 
IA-FSNC                                         & 0.772           & 0.756          & 0.793           & 0.782          & 0.699           & 0.675           & 0.729            & 0.706           & 0.650           & 0.643           & 0.675           & 0.660          & 0.565           & 0.541           & 0.604           & 0.579           \\ 
TGRM                                         & 0.779           & 0.768          & 0.801           & 0.791          & 0.725           & 0.701           & 0.727            & 0.715           & \underline{0.656}           & 0.634           & \underline{0.699}           & \underline{0.691}          & 0.592          & 0.581           & 0.616           & 0.609           \\ \hline
Meta-GPS                                         & 0.803           & \underline{0.796}          & \underline{0.826}           & \underline{0.816}          & 0.731           & \underline{0.732}           & 0.741            & \underline{0.744}           & 0.651           & \underline{0.643}           & 0.692           & 0.686          & \underline{0.612}          & \underline{0.603}           & \underline{0.642}           & \underline{0.641}           \\
Meta-GPS++                                          & \textbf{0.837}  & \textbf{0.833} & \textbf{0.849}  & \textbf{0.846} & \textbf{0.767}  & \textbf{0.751}  & \textbf{0.771}  & \textbf{0.764}  & \textbf{0.691}  & \textbf{0.683}  & \textbf{0.722}  & \textbf{0.716}  & \textbf{0.650}  & \textbf{0.643}& \textbf{0.662}  & \textbf{0.651}  \\ \hline
%\multicolumn{1}{c|}{Ab Impro}                 & 6.1\%              & 6.4\%           & 6\%            & 5.9\%           & 6.8\%            & 7.1\%            & 6.1\%             & 6.5\%            & 7.6\%            & 7.4\%           & 6.8\%            & 7\%           & 7.3\%            & 7.7\%            & 6.1\%              & 6.6\%            \\ \hline
\end{tabular}%
}

\centering
\resizebox{0.9\textwidth}{!}{%
\begin{tabular}{l|cccccccc|cccccccc}
\hline
\multicolumn{1}{c|}{\multirow{3}{*}{Models}} & \multicolumn{8}{c|}{Amazon-Electronics}                                                                                                      & \multicolumn{8}{c}{DBLP}                                                                                                                    \\ \cline{2-17} 
\multicolumn{1}{c|}{}                        & \multicolumn{2}{c}{5-way 3-shot} & \multicolumn{2}{c}{5-way 5-shot} & \multicolumn{2}{c}{10-way 3-shot} & \multicolumn{2}{c|}{10-way 5-shot} & \multicolumn{2}{c}{5-way 3-shot} & \multicolumn{2}{c}{5-way 5-shot} & \multicolumn{2}{c}{10-way 3-shot} & \multicolumn{2}{c}{10-way 5-shot} \\ \cline{2-17} 
\multicolumn{1}{c|}{}                        & ACC             & F1             & ACC             & F1             & ACC             & F1              & ACC              & F1              & ACC             & F1             & ACC             & F1             & ACC             & F1              & ACC             & F1              \\ \hline
DeepWalk                                     & 0.235           & 0.222          & 0.261           & 0.257          & 0.147           & 0.129           & 0.160            & 0.147           & 0.447           & 0.431          & 0.624           & 0.604          & 0.338           & 0.308           & 0.451           & 0.430           \\
node2vec                                     & 0.255           & 0.237          & 0.271           & 0.243          & 0.151           & 0.131           & 0.177            & 0.155           & 0.407           & 0.385          & 0.586           & 0.572          & 0.315           & 0.278           & 0.412           & 0.396           \\ \hline
GCN                                          & 0.538           & 0.498          & 0.596           & 0.553          & 0.423           & 0.384           & 0.474            & 0.483           & 0.596           & 0.549          & 0.683           & 0.660           & 0.439           & 0.390           & 0.512           & 0.476           \\
SGC                                          & 0.546           & 0.534          & 0.608           & 0.594          & 0.432           & 0.415           & 0.500            & 0.476           & 0.573           & 0.547          & 0.650           & 0.621          & 0.402           & 0.368           & 0.503           & 0.464           \\ \hline
Protonet                                     & 0.535           & 0.556          & 0.597           & 0.615          & 0.399           & 0.400           & 0.450            & 0.448           & 0.372           & 0.367          & 0.434           & 0.443          & 0.262           & 0.260           & 0.326           & 0.328           \\
MAML                                         & 0.533           & 0.521          & 0.590           & 0.583          & 0.374           & 0.361           & 0.434            & 0.413           & 0.397           & 0.397          & 0.455           & 0.437          & 0.308           & 0.253           & 0.347           & 0.312           \\ \hline
Meta-GNN                                     & 0.632           & 0.615          & 0.679           & 0.668          & 0.582           & 0.558           & 0.608            & 0.601           & 0.709           & 0.703          & 0.742           & 0.744          & 0.607           & 0.604           & 0.631           & 0.622           \\
GPN                                          & 0.646           & 0.628          & 0.709           & 0.706          & 0.603           & 0.607           & 0.624            & 0.637           & 0.745           & 0.739          & 0.761           & 0.758          & 0.626           & 0.626           & 0.640           & 0.644           \\
G-Meta                                       & 0.695     & 0.691    & 0.736     & {0.726}    & {0.664}     & {0.646}     & 0.680      & 0.672     & 0.751     & 0.743    & 0.776     & 0.769    & 0.632     & 0.636     & 0.642     & 0.646     \\
TENT                                         & 0.758           & 0.756          & 0.794           & 0.775          & 0.676           & 0.651           & 0.698            & 0.673           & 0.790           & 0.781           & 0.828           & 0.813          & 0.655           & 0.641           & 0.724           & 0.710           \\ 
IA-FSNC                                         & 0.725           & 0.708          & 0.776           & 0.750          & 0.663           & 0.649           & 0.692            & 0.669
& 0.801           & 0.786           & 0.832           & 0.813          & 0.682           & 0.665           & 0.704           & 0.696           \\ 
TGRM                                         & 0.761           & 0.748          & 0.816           & 0.781          & 0.745           & 0.729           & 0.779            & 0.756           & 0.817           & 0.803           & 0.843           & 0.831          & 0.712           & 0.691           & 0.734           & 0.720           \\ \hline
Meta-GPS                                         & \underline{0.828}           & \underline{0.827}          & \underline{0.852}           & \underline{0.830}          & \underline{0.785}           & \underline{0.782}           & \underline{0.793}            & \underline{0.791}           & \underline{0.849}           & \underline{0.847}           & \underline{0.859}           & \underline{0.855}          & \underline{0.759}           & \underline{0.751}           & \underline{0.779}           & \underline{0.772}           \\
Meta-GPS++                                          & \textbf{0.839}  & \textbf{0.833} &  \textbf{0.869}  & \textbf{0.847} &  \textbf{0.805}  & \textbf{0.791}  & \textbf{0.813}   & \textbf{0.802}  & \textbf{0.858}  & \textbf{0.857} & \textbf{0.869}  & \textbf{0.862} & \textbf{0.772}  & \textbf{0.764}  & \textbf{0.783}  & \textbf{0.779} \\ \hline
%\multicolumn{1}{c|}{Ab Impro}                 & 13.3\%              & 13.6\%           & 11.6\%            & 10.4\%           & 12.1\%            & 13.6\%            & 11.3\%             & 11.9\%            & 9.8\%            & 10.4\%           & 9.3\%            & 8.6\%           & 12.7\%            & 11.5\%            & 11.7\%              & 12.6\%            \\ \hline
\end{tabular}%
}

\label{homophily}
\end{table*}

\begin{table*}[ht]
\centering
\caption{Test accuracy on different datasets with various noisy data ratios. $\ast$: out of memory.}
\label{label noise}
\resizebox{0.9\textwidth}{!}{%
\begin{tabular}{l|cccccc|cccccc}
\hline
\multicolumn{1}{c|}{\multirow{3}{*}{Models}} &
  \multicolumn{6}{c|}{Reddit} &
  \multicolumn{6}{c}{ogbn-products} \\ \cline{2-13} 
\multicolumn{1}{c|}{} & \multicolumn{3}{c|}{5-way 5-shot} & \multicolumn{3}{c|}{10-way 5-shot} & \multicolumn{3}{c|}{5-way 5-shot} & \multicolumn{3}{c}{10-way 5-shot} \\ \cline{2-13} 
\multicolumn{1}{c|}{} &
  10\% &
  20\% &
  \multicolumn{1}{c|}{30\%} &
  10\% &
  20\% &
  30\% &
  10\% &
  20\% &
  \multicolumn{1}{c|}{30\%} &
  10\% &
  20\% &
  30\% \\ \hline
Meta-GNN &
  0.603 &
  0.576 &
  \multicolumn{1}{c|}{0.553} &
  0.490 &
  0.465 &
  0.439 &
  0.496 &
  0.472 &
  \multicolumn{1}{c|}{0.442} &
  0.369 &
  0.346 &
  0.325 \\
GPN &
  0.662 &
  0.637 &
  \multicolumn{1}{c|}{0.611} &
  0.556 &
  0.531 &
  0.502 &
  0.551 &
  0.532 &
  \multicolumn{1}{c|}{0.507} &
  0.402 &
  0.389 &
  0.352 \\
G-Meta &
  0.713 &
  0.693 &
  \multicolumn{1}{c|}{0.670} &
  0.605 &
  0.582 &
  0.556 &
  0.607 &
  0.586 &
  \multicolumn{1}{c|}{0.556} &
  0.459 &
  0.432 &
  0.401 \\
TENT &
  $\ast$ &
  $\ast$ &
  \multicolumn{1}{c|}{$\ast$} &
  $\ast$ &
  $\ast$ &
  $\ast$ &
  $\ast$ &
  $\ast$ &
  \multicolumn{1}{c|}{$\ast$} &
  $\ast$ &
  $\ast$ &
  $\ast$ \\
IA-FSNC &
  0.719 &
  0.692 &
  \multicolumn{1}{c|}{0.669} &
  0.615 &
  0.589 &
  0.562 &
  0.622 &
  0.603 &
  \multicolumn{1}{c|}{0.572} &
  0.486 &
  0.451 &
  0.419 \\
TGRM &
  0.732 &
  0.716 &
  \multicolumn{1}{c|}{0.690} &
  0.616 &
  0.591 &
  0.569 &
  0.656 &
  0.636 &
  \multicolumn{1}{c|}{0.597} &
  0.519 &
  0.492 &
  0.465 \\
Meta-GPS &
  0.771 &
  0.759 &
  \multicolumn{1}{c|}{0.742} &
  0.646 &
  0.635 &
  0.619 &
  0.655 &
  0.642 &
  \multicolumn{1}{c|}{0.616} &
  0.522 &
  0.509 &
  0.486 \\
Meta-GPS++ &
  \textbf{0.807} &
  \textbf{0.792} &
  \multicolumn{1}{c|}{\textbf{0.776}} &
  \textbf{0.676} &
  \textbf{0.669} &
  \textbf{0.653} &
  \textbf{0.689} &
  \textbf{0.682} &
  \multicolumn{1}{c|}{\textbf{0.659}} &
  \textbf{0.549} &
  \textbf{0.538} &
  \textbf{0.516} \\ \hline
\end{tabular}%
}
%\end{table*}

\centering
\resizebox{0.9\textwidth}{!}{%
\begin{tabular}{l|cccccc|cccccc}
\hline
\multicolumn{1}{c|}{\multirow{3}{*}{Models}} &
  \multicolumn{6}{c|}{Amazon-Clothing} &
  \multicolumn{6}{c}{Cora-Full} \\ \cline{2-13} 
\multicolumn{1}{c|}{} & \multicolumn{3}{c|}{5-way 5-shot} & \multicolumn{3}{c|}{10-way 5-shot} & \multicolumn{3}{c|}{5-way 5-shot} & \multicolumn{3}{c}{10-way 5-shot} \\ \cline{2-13} 
\multicolumn{1}{c|}{} &
  10\% &
  20\% &
  \multicolumn{1}{c|}{30\%} &
  10\% &
  20\% &
  30\% &
  10\% &
  20\% &
  \multicolumn{1}{c|}{30\%} &
  10\% &
  20\% &
  30\% \\ \hline
Meta-GNN &
  0.747 &
  0.721 &
  \multicolumn{1}{c|}{0.692} &
  0.619 &
  0.601 &
  0.579 &
  0.570 &
  0.552 &
  \multicolumn{1}{c|}{0.522} &
  0.511 &
  0.496 &
  0.465 \\
GPN &
  0.761 &
  0.732 &
  \multicolumn{1}{c|}{0.697} &
  0.651 &
  0.625 &
  0.594 &
  0.581 &
  0.562 &
  \multicolumn{1}{c|}{0.535} &
  0.542 &
  0.523 &
  0.501 \\
G-Meta &
  0.737 &
  0.715 &
  \multicolumn{1}{c|}{0.696} &
  0.659 &
  0.637 &
  0.619 &
  0.606 &
  0.585 &
  \multicolumn{1}{c|}{0.556} &
  0.559 &
  0.535 &
  0.516 \\
TENT &
  0.790 &
  0.761 &
  \multicolumn{1}{c|}{0.722} &
  0.720 &
  0.691 &
  0.652 &
  0.671 &
  0.660 &
  \multicolumn{1}{c|}{0.639} &
  0.549 &
  0.526 &
  0.502 \\
IA-FSNC &
  0.747 &
  0.721 &
  \multicolumn{1}{c|}{0.692} &
  0.619 &
  0.601 &
  0.579 &
  0.656 &
  0.630 &
  \multicolumn{1}{c|}{0.602} &
  0.582 &
  0.569 &
  0.536 \\
TGRM &
  0.781 &
  0.759 &
  \multicolumn{1}{c|}{0.732} &
  0.710 &
  0.691 &
  0.660 &
  0.682 &
  0.663 &
  \multicolumn{1}{c|}{0.632} &
  0.589 &
  0.552 &
  0.525 \\
Meta-GPS &
  0.819 &
  0.807 &
  \multicolumn{1}{c|}{0.792} &
  0.736 &
  0.725 &
  0.710 &
  0.675 &
  0.662 &
  \multicolumn{1}{c|}{0.646} &
  0.622 &
  0.599 &
  0.566 \\
Meta-GPS++ &
  \textbf{0.829}&
  \textbf{0.818} &
  \multicolumn{1}{c|}{\textbf{0.806}} &
  \textbf{0.759} &
  \textbf{0.749} &
  \textbf{0.730} &
  \textbf{0.703} &
  \textbf{0.695} &
  \multicolumn{1}{c|}{\textbf{0.679}} &
  \textbf{0.649} &
  \textbf{0.628} &
  \textbf{0.619} \\ \hline
\end{tabular}%
}

\centering
\resizebox{0.9\textwidth}{!}{%
\begin{tabular}{l|cccccc|cccccc}
\hline
\multicolumn{1}{c|}{\multirow{3}{*}{Models}} &
  \multicolumn{6}{c|}{Amazon-Electronics} &
  \multicolumn{6}{c}{DBLP} \\ \cline{2-13} 
\multicolumn{1}{c|}{} & \multicolumn{3}{c|}{5-way 5-shot} & \multicolumn{3}{c|}{10-way 5-shot} & \multicolumn{3}{c|}{5-way 5-shot} & \multicolumn{3}{c}{10-way 5-shot} \\ \cline{2-13} 
\multicolumn{1}{c|}{} &
  10\% &
  20\% &
  \multicolumn{1}{c|}{30\%} &
  10\% &
  20\% &
  30\% &
  10\% &
  20\% &
  \multicolumn{1}{c|}{30\%} &
  10\% &
  20\% &
  30\% \\ \hline
Meta-GNN &
  0.647 &
  0.615 &
  \multicolumn{1}{c|}{0.586} &
  0.590 &
  0.566 &
  0.542 &
  0.735 &
  0.719 &
  \multicolumn{1}{c|}{0.696} &
  0.611 &
  0.596 &
  0.575 \\
GPN &
  0.682 &
  0.651 &
  \multicolumn{1}{c|}{0.633} &
  0.610 &
  0.601 &
  0.576 &
  0.758 &
  0.742 &
  \multicolumn{1}{c|}{0.723} &
  0.626 &
  0.613 &
  0.591 \\
G-Meta &
  0.714 &
  0.697 &
  \multicolumn{1}{c|}{0.675} &
  0.661 &
  0.633 &
  0.606 &
  0.761 &
  0.745 &
  \multicolumn{1}{c|}{0.722} &
  0.630 &
  0.619 &
  0.596 \\
TENT &
  0.769 &
  0.747 &
  \multicolumn{1}{c|}{0.715} &
  0.678 &
  0.655 &
  0.625 &
  0.805 &
  0.789 &
  \multicolumn{1}{c|}{0.773} &
  0.704 &
  0.692 &
  0.665 \\
IA-FSNC &
  0.756 &
  0.740 &
  \multicolumn{1}{c|}{0.722} &
  0.682 &
  0.671 &
  0.656 &
  0.802 &
  0.780 &
  \multicolumn{1}{c|}{0.762} &
  0.682 &
  0.671 &
  0.653 \\
TGRM &
  0.795 &
  0.773 &
  \multicolumn{1}{c|}{0.745} &
  0.760 &
  0.736 &
  0.712 &
  0.812 &
  0.796 &
  \multicolumn{1}{c|}{0.775} &
  0.719 &
  0.692 &
  0.669 \\
Meta-GPS &
  0.841 &
  0.829 &
  \multicolumn{1}{c|}{0.814} &
  0.786 &
  0.775 &
  0.759 &
  0.835 &
  0.826 &
  \multicolumn{1}{c|}{0.812} &
  0.762 &
  0.739 &
  0.726 \\
Meta-GPS++ &
  \textbf{0.856} &
  \textbf{0.840} &
  \multicolumn{1}{c|}{\textbf{0.822}} &
  \textbf{0.802} &
  \textbf{0.790} &
  \textbf{0.776} &
  \textbf{0.853} &
  \textbf{0.845} &
  \multicolumn{1}{c|}{\textbf{0.832}} &
  \textbf{0.775} &
  \textbf{0.762} &
  \textbf{0.749} \\ \hline
\end{tabular}%
}
\end{table*}

\subsection{Model Implementation}
In the \textit{graph network encoder}, we use 2-hop neighborhoods, \textit{i.e.}, $\ell=2$ in Eq. \ref{esgc} and set the dimension of the final learned embedding $d^{\prime}$ to 16. In Eqs. \ref{init} and \ref{s2}, we uniformly use the MLP architecture with a 16-dimensional hidden layer activated by the ReLU function and a linear output layer. We set the step size $\alpha$ and meta-learning rate $\beta$ to 0.5 and 0.001, respectively. Moreover, the regularization coefficients $\xi,\zeta,\gamma$ are set to 0.1, 0.1, and 0.001. The temperature $\tau$ in Eq. \ref{cl} is 0.5 and K in Eq. \ref{topk} is 30. The number of tasks per batch and the size $M$ of the query set in each task are uniformly set to 10. In addition, we perform gradient descent updates uniformly for five steps during both the meta-training and meta-testing stages. To prevent overfitting, we implement an early stopping strategy based on validation performance. Specifically, if the validation loss does not decrease for 50 consecutive epochs, we stop the model training. All experiments are run on a 24 GB Nvidia GeForce 3090Ti GPU. 
%\textbf{Evaluation Metric.} 
%\subsection{Evaluation Metric}
To evaluate the performance of the model, we use Accuracy (ACC) and Macro-F1 score (F1) which are widely used in previous models for few-shot node classification on graphs. 
% The mathematical expression for the evaluation metrics can be expressed as:
% \begin{equation}
%     \begin{aligned}
%         \text{Accuracy} &= \frac{\# \text{correct predictions}}{\# \text{total predictions}}, \quad
%         \text{Micro-F1} = \frac{2\times \text{Micro-pre}\times \text{Micro-rec}}{\text{Micro-pre}+\text{Micro-rec}} \\
%         \text{Micro-pre} &=\frac{\sum_{c\in C}\text{TP}_c}{\sum_{c\in C}(\text{TP}_c+\text{FP}_c)}, \quad
%         \text{Micro-rec} =\frac{\sum_{c\in C}\text{TP}_c}{\sum_{c\in C}(\text{TP}_c+\text{FN}_c)}
%     \end{aligned}
% \end{equation}
% where TP$_c$, FP$_c$, and FN$_c$ represent the number of true positives, false positives, and false negatives for the class $c$, respectively.
%The evaluation metrics in our experiments are Accuracy (ACC) and Micro-F1 score (F1), which are widely used in few-shot node classification on attributed networks.
\section{Results}
%\textbf{Model Performance.} 
\subsection{Evaluation of Model Performance}
To verify the effectiveness of our proposed model, we conduct thorough experiments on Meta-GPS++ and other baseline models with four experimental settings for each dataset. Note that Meta-GPS is the model proposed in our previous conference version. To ensure the fairness and stability of the experiments, we randomly select 200 meta-testing tasks from $\mathcal{Y}_{test}$ to evaluate the performance of each model. This process is repeated ten times to obtain a reliable average of experimental results, which are presented in Table \ref{homophily}. Based on these quantitative results, we provide a detailed analysis of our observations and findings.

\noindent $\bullet$ Under different experimental settings, our model clearly achieves the best results compared to other baseline models for all datasets. For example, in the 5-way 5-shot experimental setting on the Reddit dataset, our proposed model Meta-GPS++ gains an absolute improvement of 4.3\% and 3.8\% in accuracy and F1 score over our previous model Meta-GPS, and an absolute improvement of 7.1\% and 7.8\% over the second best performing model TGRM. This demonstrates the designed model's superiority and strong generalization ability. We attribute this performance to the following design choices. First, to alleviate the adverse effects of stochasticity in the MAML learning process, we introduce prototype-based parameter initialization and contrastive learning for task randomness. The former can generate category guidance information and alleviate the problem of model dependence on support data, thereby enhancing the robustness of the model to data noise, while the latter can regularize node embedding distribution by contrastive learning. Second, we adopt self-training techniques to effectively utilize the rich self-supervised and task-relevant information contained in the unselected nodes in the graph that are previously ignored by other models. This can regularize training of the meta-model and alleviate overfitting problems. Third, we incorporate scaling and shifting vectors, which enable customized parameterization during training, enhancing the efficiency of transferable knowledge extraction on unseen tasks.%Overall, our model achieves significantly superior performance on all datasets under different experimental settings compared to other competitive baselines, illustrating its powerful generalization capability. The reason is that our model produces prototype-based parameter initialization instead of a single random initialization, which provides category-guidance information, making it possible to alleviate the reliance on support data and to enhance the robustness of the model to outliers. We also employ scaling and shifting vectors to provide customized trainable parameters for training tasks; this allows for a more efficient extraction of transferable knowledge for adaptation to new tasks.

\noindent $\bullet$ We find that our proposed Meta-GPS++ outperforms other competitive baseline models in both homophilic and heterophilic graphs. For example, taking the accuracy for 10-way 3-shot learning on three datasets, Reddit, Amazon-Clothing, and Amazon-Electronics as an example, compared with the second-best model TGRM, Meta-GPS++ achieves absolute improvements of 4.3\%, 4.2\%, and 6\%, respectively. This may be attributed to our specially designed graph layer for effectively learning discriminative node embeddings for different classes and adopting non-aggregated representations rather than mixing ego-nodes and their neighbors to learn node embeddings, thus reducing the noise when learning node features in graphs, especially in heterophilic graphs. %We find that Meta-GPS achieves the best performance in both homophilic and heterophilic attributed networks, while other baselines generally perform relatively poorly on heterophilic datasets. An interesting finding is that as the homophily level of a network decreases, our model achieves higher absolute improvements compared to other baselines. Taking the example of our model's performance on the task of 5-way 3-shot in Reddit, Cora-Full and Amazon-Electronics, the model achieves 4\%, 7.6\% and 13.3\% absolute improvements in Accuracy, respectively, compared to that of the best-performing baseline G-Meta. We attribute this to the use of a specifically designed graph layer for learning more expressive node representations in a non-aggregated manner. When a network exhibits high homophily, both our designed graph layer and classic graph layers can extract discriminative features of different node classes. In contrast, when the network is highly heterophilic, other models still cumulatively learn node representations, which introduces more noise to the learned node representations and is not beneficial for learning the essential patterns of different classes.

\noindent $\bullet$ We observe that graph meta-learning models (such as Meta-GNN, GPN, and G-Meta) outperform the other three types of baseline models. This is because these models simultaneously consider both node features and topological information of graphs and adopt the paradigm of meta-learning for training. Taking 5-way 5-shot learning on the Reddit dataset as an example, G-Meta outperforms SGC, which performs the best among the first two types of models by 26.7\% in terms of accuracy. Moreover, some recently proposed graph meta-learning models (TENT, IA-FSNC, and TGRM) have further achieved promising performance in these datasets due to integrating advanced technologies. %IA-FSNC adopts both support augmentation and shot augmentation during graph meta-learning, while TGRM introduces graph contrastive learning into graph meta-learning by considering the relationship between tasks. %Random walk-based models (\textit{e.g.} DeepWalk, node2vec) and GNN-based models (\textit{e.g.} GCN, SGC) achieve unsatisfactory performance in FSL scenarios, lagging far behind graph meta-learning models. A possible reason is that these models typically require large quantities of labeled data to obtain competitive results and suffer from severe overfitting problems when only a small number of labeled nodes are available. Notably, despite the fruitful success of traditional meta-learning methods (\textit{e.g.} Protonet, MAML) in few-shot image classification, these methods perform worse than the GNN-based models in our few-shot node classification scenarios. This can be attributed to the fact that they neglect the graph topology information of attributed networks, which adversely affects the learned node representation. Since simultaneously considering the meta-learning and graph topology, graph meta-learning models (\textit{e.g.} Meta-GNN, GPN, G-Meta) obtain significant gains compared to the other two types of models. However, due to the limitations mentioned in the introduction, their performance is still far behind Meta-GPS.

\noindent $\bullet$ In many experimental settings, random walk-based models, such as DeepWalk and node2vec, perform far worse than other models in few-shot node classification. One possible reason is that they only focus on learning good node embeddings in unsupervised scenarios, while training classifiers with only a few labeled data encounters serious overfitting problems. Moreover, GNN-based models (such as GCN and SGC) also do not perform well in the studied scenarios. Although they consider label information when learning node embeddings, there are too few labeled nodes in this scenario to distinguish between different categories of nodes effectively. For example, in the 5-way 3-shot few-shot setting on the Cora-Full dataset, GCN even lags behind Meta-GNN in accuracy and F1 score, reaching 17.6\% and 17.1\%, which fully validates our previous analysis.

\noindent $\bullet$ Traditional meta-learning models such as Protonet and MAML perform well in few-shot image classification scenarios but poorly in few-shot node classification. In fact, in most cases, they are even inferior to GNN-based models. For example, in the 5-way 3-shot experimental setting on the DBLP dataset, Protonet even lags behind GCN in accuracy and F1 score, reaching 22.4\% and 18.2\%, respectively. One plausible reason for this gap is that traditional models completely ignore graph structure information and only treat data as independent and identically distributed text or images, which hinders node representation learning.

%\textbf{Robustness to Noisy Data.} 
\subsection{Robustness to Noisy Data}
As stated in the introduction, current MAML-based methods heavily depend on instance-based statistics, making their performance highly vulnerable to data noise. To further validate the resilience of Meta-GPS++ when the support instances contain noise, we randomly transform a specific proportion of the support instances from each class into noisy instances that are sampled from other classes, in line with \cite{dong2020meta}. Then, each class within a meta-task can be regarded to have a certain percentage of noise, given that we sample numerous meta-tasks for training. According to the results presented in Table \ref{label noise}, we can clearly observe that Meta-GPS++ achieves the best performance under various noise data ratios across different datasets compared to other competitive baselines, which illustrates the effectiveness of the model design. Moreover, Table \ref{label noise} also shows that as the proportion of noisy data increases, Meta-GPS++ gains more benefits than other methods, indicating its excellent robustness and noise-reduction. Taking the 5-way 5-shot experimental setting on the ogbn-products dataset for example, our model Meta-GPS++ outperforms the previously proposed model Meta-GPS by 3.4\%, 4\%, and 4.3\% in the 5\%, 10\%, and 20\% noise data ratios, respectively, and outperforms the best baseline model TGRM by 3.3\%, 4.6\%, and 6.2\%, respectively. One explainable reason is that we introduce prototype-based parameter initialization and contrastive learning containing prototypes, which can help the model alleviate the adverse effects of data noise and prevent node embeddings from deviating too far from prototypes in feature space. Other models show varying degrees of performance degradation when subjected to data noise interference, indicating that most existing graph meta-models are susceptible to the influence of noisy data. In contrast, IA-FSNC and TGRM have a lighter performance decline. One possible reason is the introduction of data augmentation during meta-learning, while another may be due to the introduction of graph contrastive learning. Both techniques can alleviate the impact of data noise and promote model performance.

\begin{figure*}[ht]
    \centering
    \subfigure[5-way $K$-shot]{\includegraphics[width=0.45\textwidth]{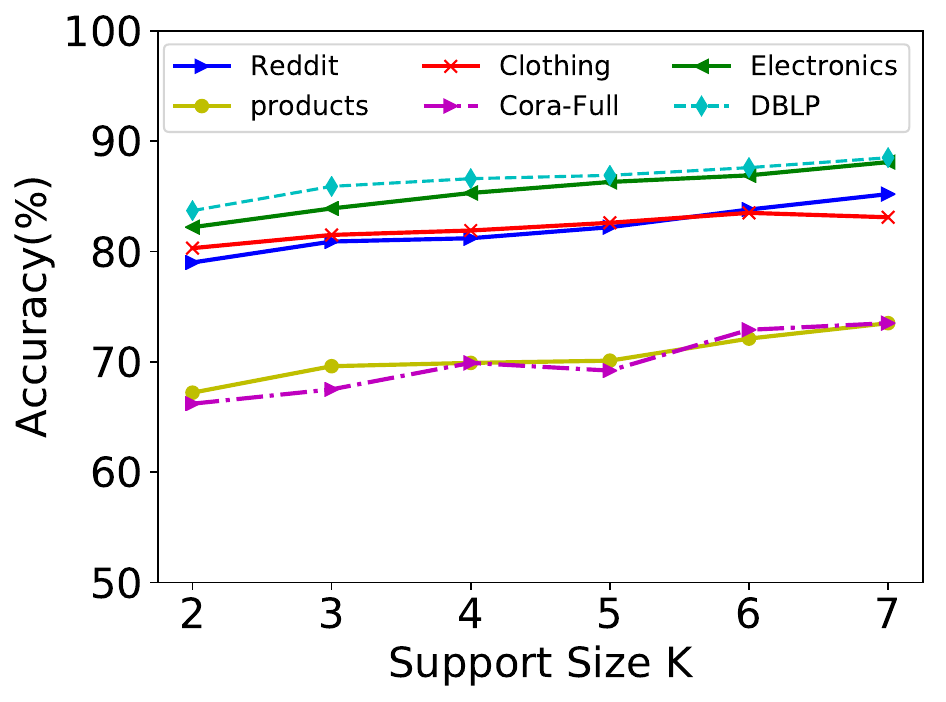}}
    \subfigure[$N$-way 5-shot]{\includegraphics[width=0.45\textwidth]{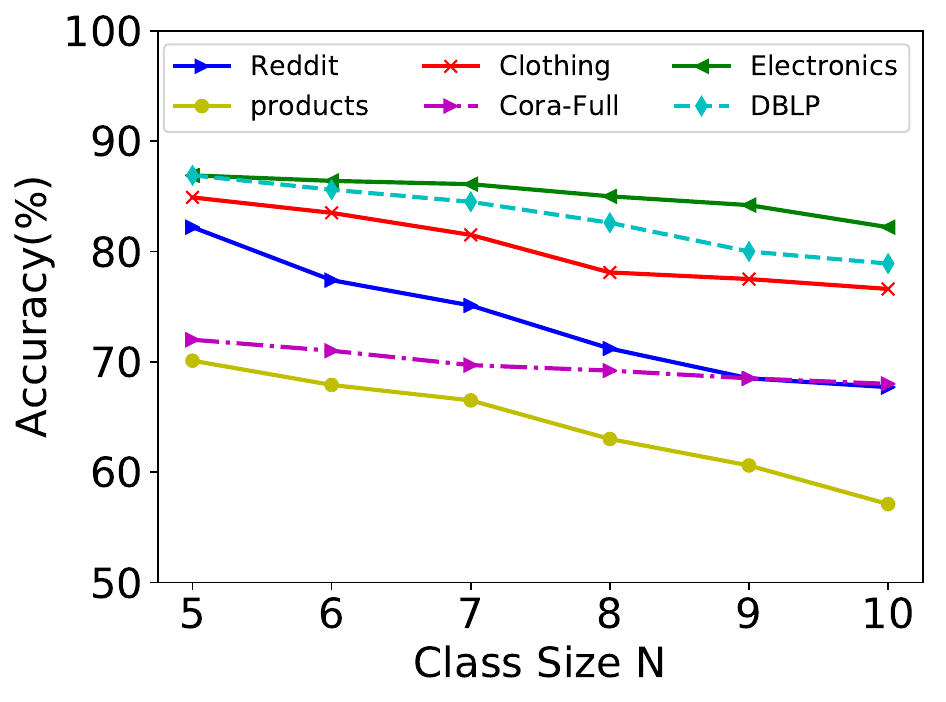}}
    \subfigure[5-way 5-shot]{\includegraphics[width=0.45\textwidth]{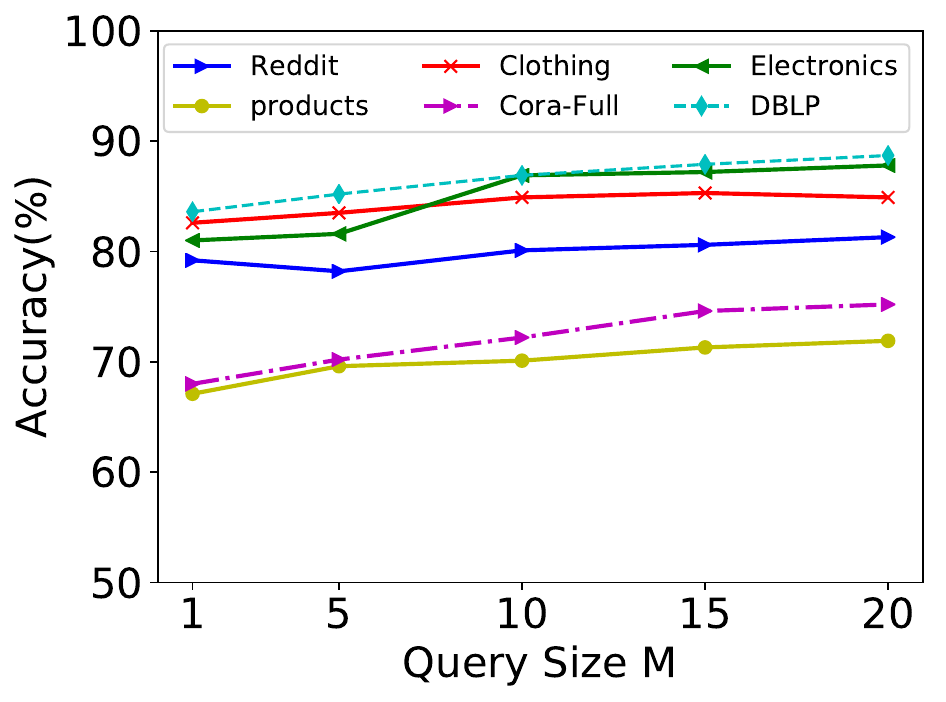}}
    \subfigure[5-way 5-shot]{\includegraphics[width=0.45\textwidth]{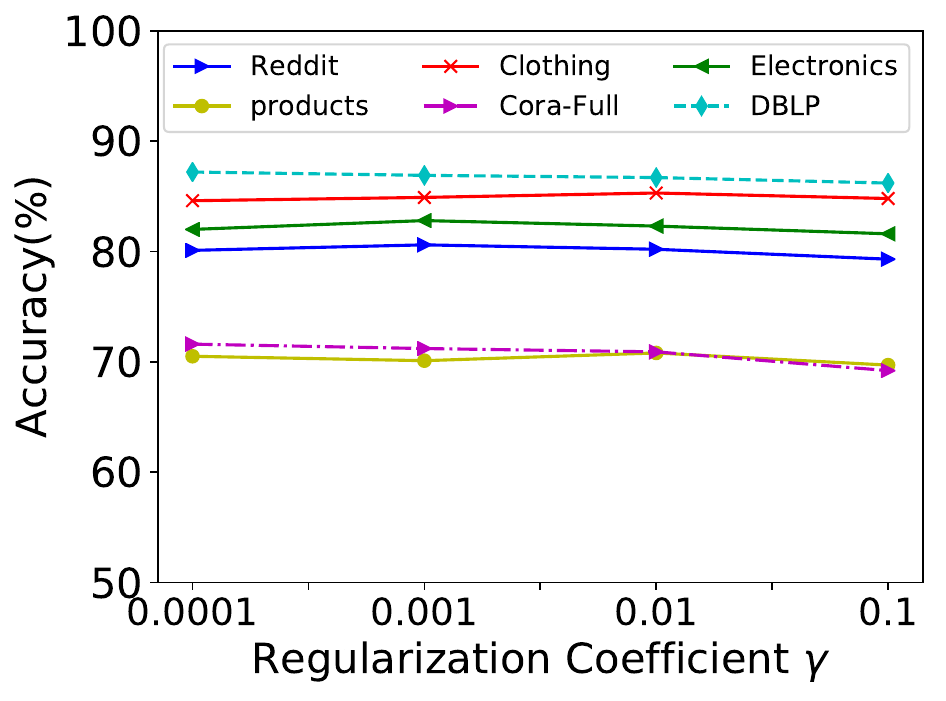}}
    \caption{Evaluation performance of Meta-GPS++ for the first group hyperparameters.}
    \label{hyperparameter_one}
\end{figure*}

\begin{figure*}[ht]
    \subfigure[5-way 5-shot]{\includegraphics[width=0.45\textwidth]{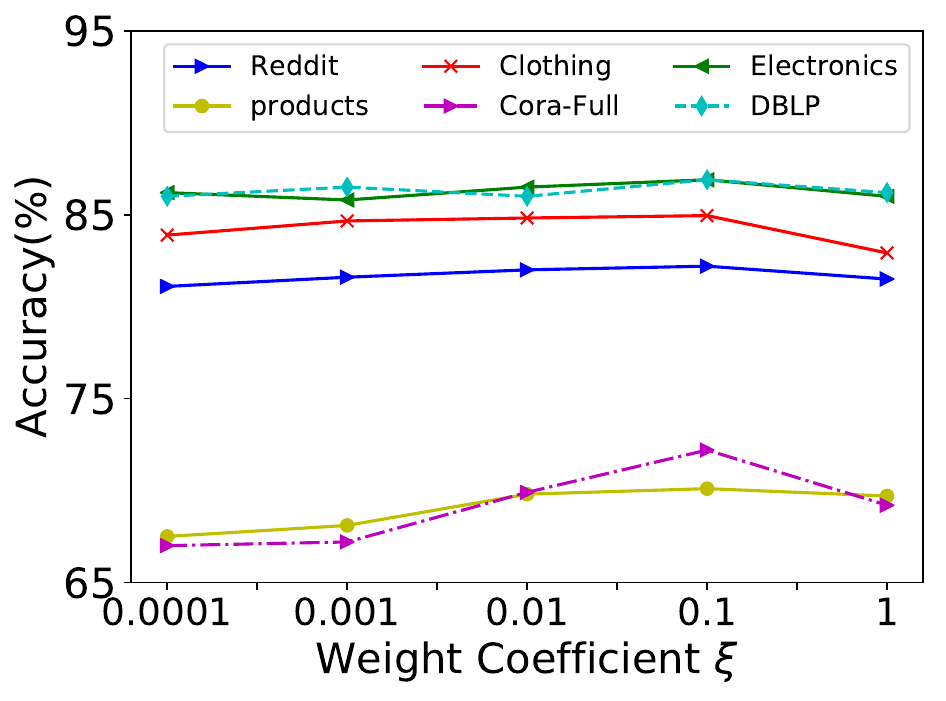}}
    \subfigure[5-way 5-shot]{\includegraphics[width=0.45\textwidth]{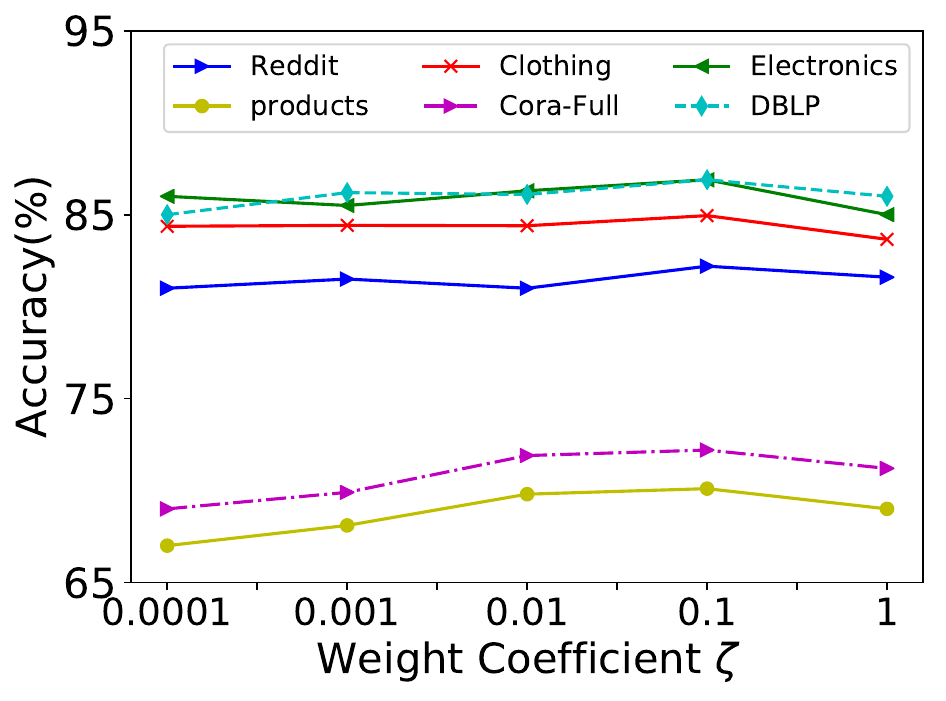}}
    \subfigure[5-way 5-shot]{\includegraphics[width=0.45\textwidth]{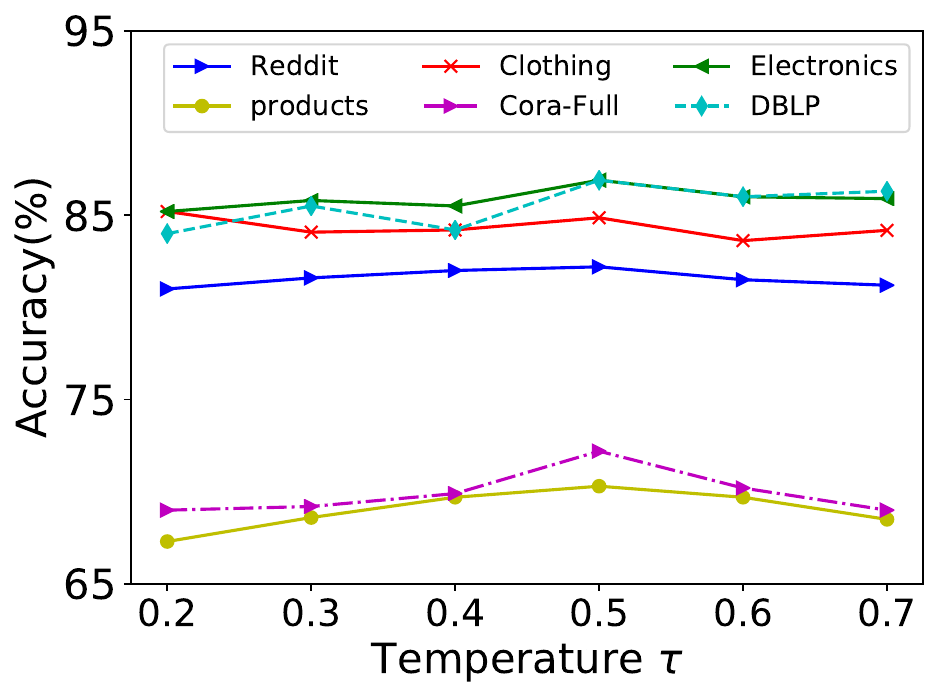}}
    \subfigure[5-way 5-shot]{\includegraphics[width=0.45\textwidth]{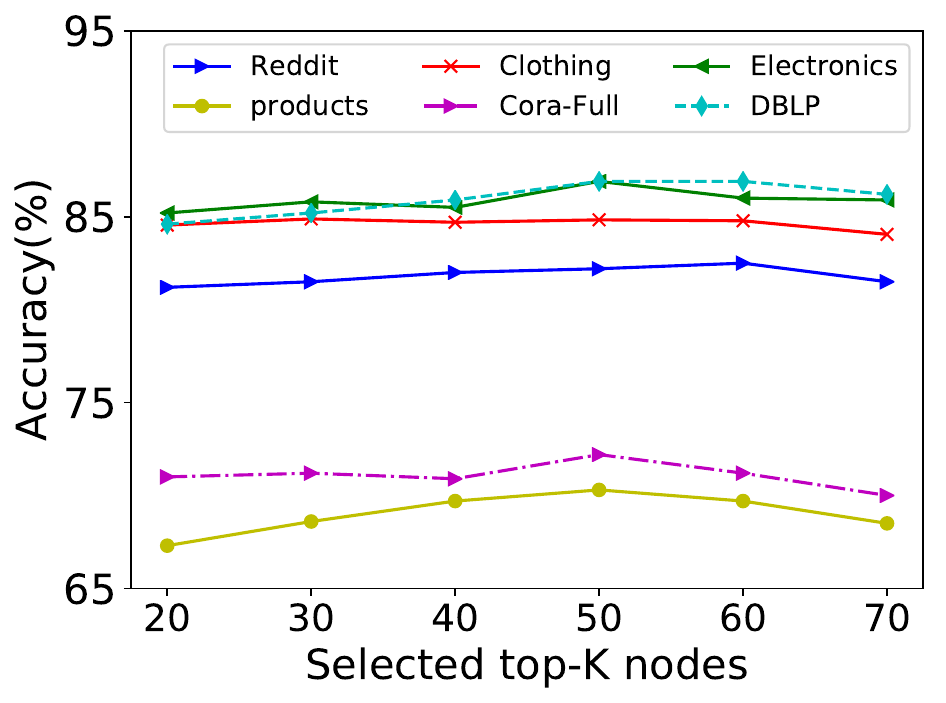}}
    \caption{Evaluation performance of Meta-GPS++ for the second group hyperparameters.}
    \label{hyperparameter_two}
\end{figure*}

% \begin{figure*}[ht]
%     \centering
% \end{figure*}

\begin{figure}[ht]
    \centering
    \subfigure[Meta-GPS++ on Amazon-Electronics]{\includegraphics[width=0.35\textwidth]{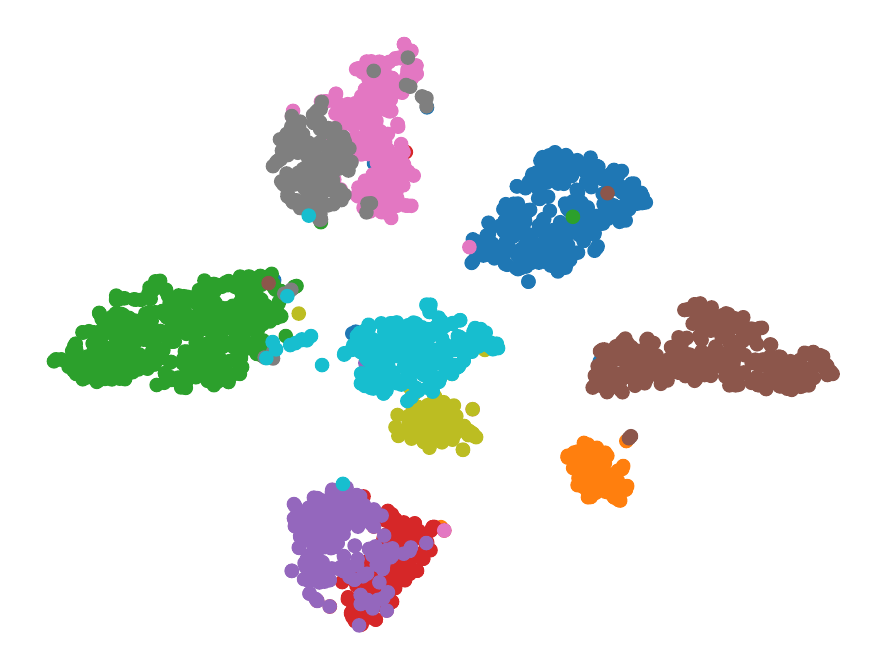}}
    \subfigure[Meta-GPS++/SGC on Amazon-Electronics]{\includegraphics[width=0.35\textwidth]{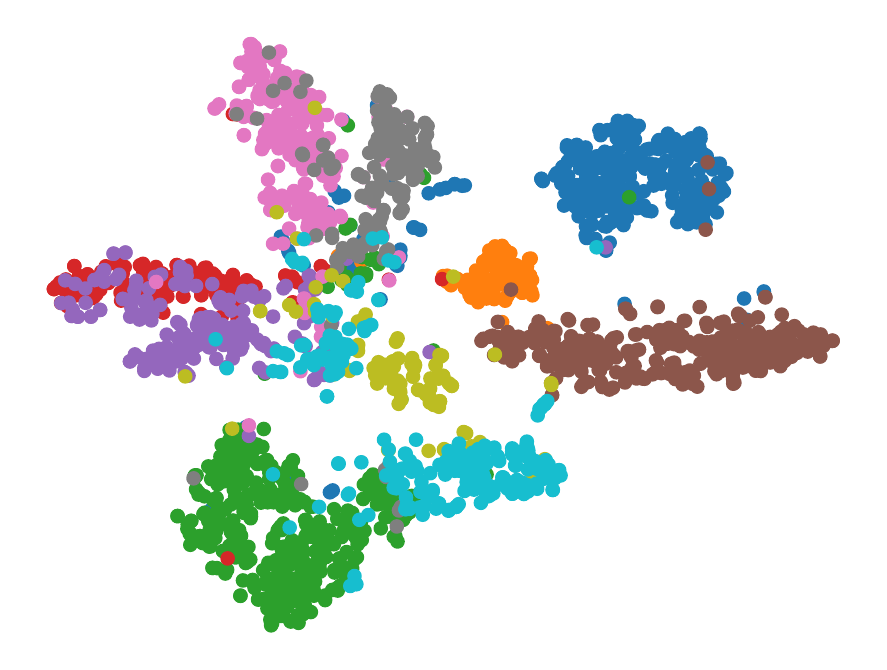}}
    \subfigure[Meta-GPS++ on DBLP]{\includegraphics[width=0.35\textwidth]{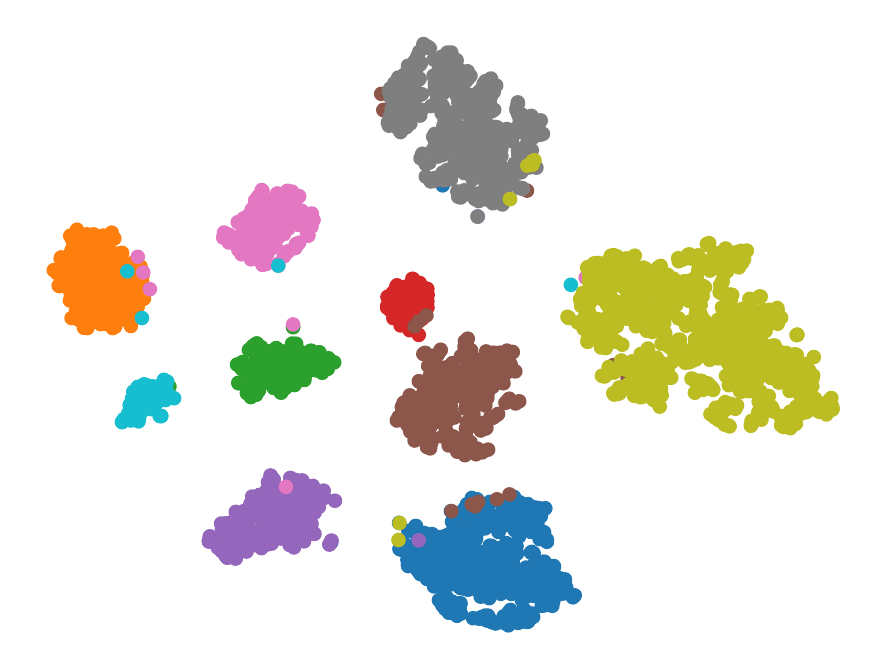}}
    \subfigure[Meta-GPS++/SGC on DBLP]{\includegraphics[width=0.35\textwidth]{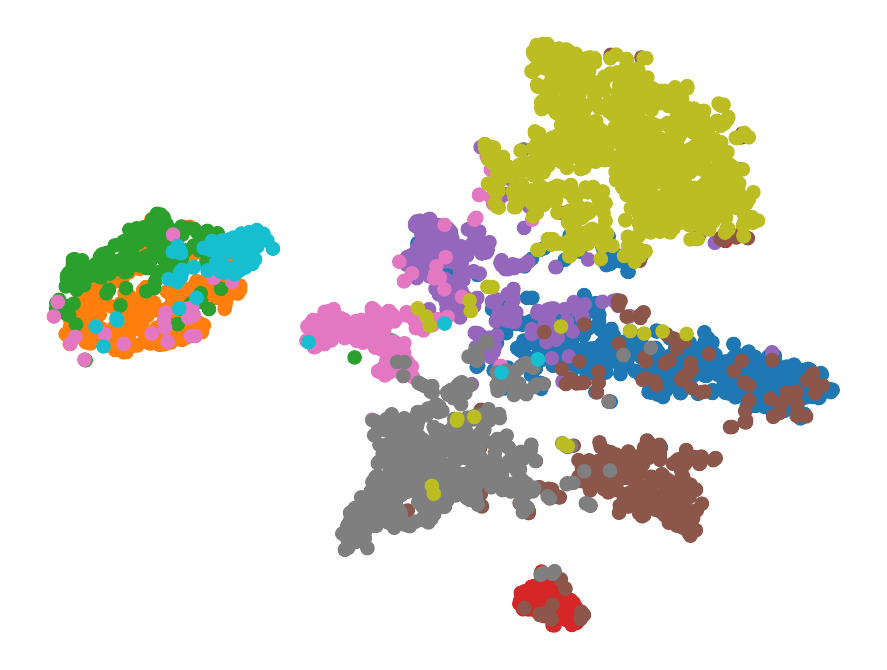}}
    \caption{Visualization of node embeddings by different methods on partial classes of test sets of Amazon-Electronics and DBLP datasets.}
    \label{tsne}
\end{figure}

%\textbf{Hyperparameter Sensitivity.} 
\subsection{Hyperparameter Sensitivity}
We examine the impact of various hyperparameter choices on our model's performance by evaluating it on all datasets. Specifically, we investigate two primary groups of hyperparameters adopted in Meta-GPS++. The first group includes the node class size ($N$-way), the support size ($K$-shot), the query set size $M$, and the regularization coefficient $\gamma$ in Eq. \ref{loss}. The second group comprises the contrastive learning coefficient $\xi$ in Eq. \ref{loss}, the self-training coefficient $\zeta$ in Eq. \ref{loss}, the temperature $\tau$ of contrastive learning in Eq. \ref{cl}, and the number of K of selected high-confidence nodes per class in Eq. \ref{topk}. The test accuracy results with different hyperparameters are presented in Figs. \ref{hyperparameter_one} and \ref{hyperparameter_two}, respectively. 

From Fig. \ref{hyperparameter_one} (a), we can observe that as the support size $K$ increases, the model's performance shows an upward trend, which is consistent with our expectations. This is because a larger support size allows the model to extract more transferable meta-knowledge and to reduce sensitivity to potentially noisy data. Conversely, as the node category $N$ increases, the model's performance clearly exhibits a downward trend, which is in accordance with our intuition. The main reason for this phenomenon is that the more test classes, the more node classes that need to be predicted, which introduces more uncertainty to the model, and thus increase the difficulty of classifying few-shot nodes. Additionally, as the query size $M$ increases, from a comprehensive view, the model's performance also improves, depicted in Fig. \ref{hyperparameter_one} (c), since it can better adapt to knowledge from larger query sets $M$ in meta-training tasks and further yield better generalization ability for the target task. Furthermore, as depicted in Fig. \ref{hyperparameter_one} (d), our model exhibits superior performance with smaller $\gamma$ values. A high value of $\gamma$ results in minimal scaling and shifting, thereby diminishing the impact of the S$^2$ transformation and causing a decline in performance. %We further empirically exploit the effects of the choices of different hyperparameters by observing the performance of our model on all evaluated datasets. Specifically, the primary hyperparameters used in Meta-GPS are the node class size ($N$-way) and the regularization coefficient $\gamma$ in Eq. \ref{loss}, respectively. Fig. \ref{hyperparameter} shows the test accuracy results with different hyperparameters. The model's performance decreases with increasing node class $N$, indicating that more uncertainty is introduced when the node classes $N$ increase. Moreover, as shown in Fig. \ref{hyperparameter} (b), for smaller $\gamma$ values, our model tends to perform better. This is because the large value leads to very little scaling and shifting, significantly reduce the role of the S$^2$ transformation and thus suffers performance degradation.

From Fig \ref{hyperparameter_two} (a), we can find that for all datasets, the model performance first increases and then decreases as the contrastive learning coefficient $\xi$ increases. This is because when the coefficient $\xi$ is too small, it has a limited effect on the model. However, when $\xi$ exceeds a threshold, it leads the contrastive learning module to dominate the model training, which is not conducive to the model's generalization. Similarly, this phenomenon also appears in the impact of the self-training coefficient $\zeta$ on the model performance, as shown in Fig. \ref{hyperparameter_two} (b). It can be noticed that the model performance reaches its peak when $\xi$ and $\zeta$ are equal to 0.1. Moreover, as shown in Fig \ref{hyperparameter_two} (c), the model achieves the peak performance when the temperature $\tau$ of contrastive learning is 0.5. One possible reason is that when the $\tau$ is too small, the model focuses on hard negative samples, which makes it too sensitive and specific, thus reducing its generalization to new data. When the $\tau$ is too large, the representations of positive and negative pairs become too similar, making it difficult for the model to correctly distinguish between different classes. Fig. \ref{hyperparameter_two} (d) indicates that the optimal results can be obtained when the number of K high-confidence nodes per class is in the range [40, 60]. A reasonable reason is that a smaller number of high-confidence nodes cannot impact the data distribution, whereas a larger number reduces the quality of the nodes.

\begin{table}[ht]
\centering
\caption{Ablation results of several designed model variants on all datasets under different few-shot settings.}
\label{ablation}
\resizebox{0.9\textwidth}{!}{%
\begin{tabular}{c|cccccccc|cccccccc}
\hline
\multirow{3}{*}{Models} & \multicolumn{8}{c|}{Reddit}                                   & \multicolumn{8}{c}{ogbn-products}                             \\ \cline{2-17} 
 &
  \multicolumn{2}{c}{5-way 3-shot} &
  \multicolumn{2}{c}{5-way 5-shot} &
  \multicolumn{2}{c}{10-way 3-shot} &
  \multicolumn{2}{c|}{10-way 5-shot} &
  \multicolumn{2}{c}{5-way 3-shot} &
  \multicolumn{2}{c}{5-way 5-shot} &
  \multicolumn{2}{c}{10-way 3-shot} &
  \multicolumn{2}{c}{10-way 3-shot} \\ \cline{2-17} 
                       & ACC   & F1    & ACC   & F1    & ACC   & F1    & ACC   & F1    & ACC   & F1    & ACC   & F1    & ACC   & F1    & ACC   & F1    \\ \hline
Meta-GPS++ &
  \textbf{0.806} &
  \textbf{0.793} &
  \textbf{0.822} &
  \textbf{0.811} &
  \textbf{0.676} &
  \textbf{0.652} &
  \textbf{0.692} &
  \textbf{0.690} &
  \textbf{0.694} &
  \textbf{0.682} &
  \textbf{0.701} &
  \textbf{0.695} &
  \textbf{0.552} &
  \textbf{0.546} &
  \textbf{0.561} &
  \textbf{0.552} \\
\textit{w/o ST}                 & 0.797 & 0.783 & 0.813 & 0.805 & 0.669 & 0.645 & 0.686 & 0.679 & 0.683 & 0.670 & 0.690 & 0.683 & 0.545 & 0.532 & 0.553 & 0.541 \\
\textit{w/o $S^2$}              & 0.786 & 0.775 & 0.804 & 0.791 & 0.658 & 0.646 & 0.680 & 0.671 & 0.679 & 0.665 & 0.686 & 0.673 & 0.538 & 0.526 & 0.536 & 0.522 \\
\textit{w/o SGC}                & 0.792 & 0.779 & 0.807 & 0.795 & 0.662 & 0.643 & 0.678 & 0.665 & 0.680 & 0.673 & 0.682 & 0.669 & 0.540 & 0.529 & 0.540 & 0.535 \\
\textit{w/o CL}                 & 0.782 & 0.761 & 0.802 & 0.791 & 0.651 & 0.640 & 0.673 & 0.660 & 0.673 & 0.651 & 0.680 & 0.665 & 0.521 & 0.510 & 0.539 & 0.530 \\
\textit{w/o PI}                 & 0.756 & 0.749 & 0.773 & 0.765 & 0.629 & 0.609 & 0.646 & 0.639 & 0.634 & 0.626 & 0.654 & 0.636 & 0.509 & 0.502 & 0.518 & 0.507 \\ \hline
\end{tabular}%
}

\resizebox{0.9\textwidth}{!}{%
\begin{tabular}{c|cccccccc|cccccccc}
\hline
\multirow{3}{*}{Models} & \multicolumn{8}{c|}{Amazon-Clothing}                         & \multicolumn{8}{c}{Cora-Full}                                 \\ \cline{2-17} 
 &
  \multicolumn{2}{c}{5-way 3-shot} &
  \multicolumn{2}{c}{5-way 5-shot} &
  \multicolumn{2}{c}{10-way 3-shot} &
  \multicolumn{2}{c|}{10-way 5-shot} &
  \multicolumn{2}{c}{5-way 3-shot} &
  \multicolumn{2}{c}{5-way 5-shot} &
  \multicolumn{2}{c}{10-way 3-shot} &
  \multicolumn{2}{c}{10-way 5-shot} \\ \cline{2-17} 
                       & ACC   & F1    & ACC   & F1    & ACC   & F1    & ACC   & F1    & ACC   & F1    & ACC   & F1    & ACC   & F1    & ACC   & F1    \\ \hline
Meta-GPS++ &
  \textbf{0.837} &
  \textbf{0.833} &
  \textbf{0.849} &
  \textbf{0.846} &
  \textbf{0.767} &
  \textbf{0.751} &
  \textbf{0.771} &
  \textbf{0.764} &
  \textbf{0.691} &
  \textbf{0.683} &
  \textbf{0.722} &
  \textbf{0.716} &
  \textbf{0.650} &
  \textbf{0.643} &
  \textbf{0.662} &
  \textbf{0.651} \\
\textit{w/o ST}                 & 0.829 & 0.822 & 0.840 & 0.835 & 0.756 & 0.740 & 0.759 & 0.747 & 0.682 & 0.670 & 0.709 & 0.704 & 0.638 & 0.630 & 0.653 & 0.646 \\
\textit{w/o $S^2$}                 & 0.811 & 0.803 & 0.830 & 0.826 & 0.739 & 0.721 & 0.745 & 0.734 & 0.669 & 0.648 & 0.702 & 0.696 & 0.632 & 0.611 & 0.646 & 0.639 \\
\textit{w/o SGC}                & 0.819 & 0.806 & 0.835 & 0.832 & 0.746 & 0.732 & 0.751 & 0.744 & 0.679 & 0.662 & 0.706 & 0.699 & 0.635 & 0.623 & 0.650 & 0.640 \\
\textit{w/o CL}                 & 0.816 & 0.809 & 0.829 & 0.825 & 0.745 & 0.739 & 0.749 & 0.740 & 0.670 & 0.658 & 0.707 & 0.701 & 0.625 & 0.619 & 0.649 & 0.636 \\
\textit{w/o PI}                 & 0.785 & 0.779 & 0.802 & 0.796 & 0.722 & 0.709 & 0.726 & 0.715 & 0.656 & 0.635 & 0.679 & 0.671 & 0.606 & 0.592 & 0.621 & 0.612 \\ \hline
\end{tabular}%
}

\resizebox{0.9\textwidth}{!}{%
\begin{tabular}{c|cccccccc|cccccccc}
\hline
\multirow{3}{*}{Models} & \multicolumn{8}{c|}{Amazon-Electronics}                      & \multicolumn{8}{c}{DBLP}                                      \\ \cline{2-17} 
 &
  \multicolumn{2}{c}{5-way 3-shot} &
  \multicolumn{2}{c}{5-way 5-shot} &
  \multicolumn{2}{c}{10-way 3-shot} &
  \multicolumn{2}{c|}{10-way 5-shot} &
  \multicolumn{2}{c}{5-way 3-shot} &
  \multicolumn{2}{c}{5-way 5-shot} &
  \multicolumn{2}{c}{10-way 3-shot} &
  \multicolumn{2}{c}{10-way 5-shot} \\ \cline{2-17} 
                       & ACC   & F1    & ACC   & F1    & ACC   & F1    & ACC   & F1    & ACC   & F1    & ACC   & F1    & ACC   & F1    & ACC   & F1    \\ \hline
Meta-GPS++ &
  \textbf{0.839} &
  \textbf{0.833} &
  \textbf{0.869} &
  \textbf{0.847} &
  \textbf{0.805} &
  \textbf{0.791} &
  \textbf{0.813} &
  \textbf{0.802} &
  \textbf{0.858} &
  \textbf{0.857} &
  \textbf{0.869} &
  \textbf{0.862} &
  \textbf{0.772} &
  \textbf{0.764} &
  \textbf{0.783} &
  \textbf{0.779} \\
\textit{w/o ST}        & 0.830 & 0.821 & 0.857 & 0.850 & 0.793 & 0.785 & 0.805 & 0.797 & 0.846 & 0.839 & 0.856 & 0.850 & 0.753 & 0.742 & 0.774 & 0.770 \\
\textit{w/o $S^2$}     & 0.810 & 0.802 & 0.849 & 0.836 & 0.782 & 0.773 & 0.791 & 0.783 & 0.840 & 0.836 & 0.853 & 0.847 & 0.742 & 0.733 & 0.771 & 0.767 \\
\textit{w/o SGC}       & 0.802 & 0.791 & 0.835 & 0.830 & 0.772 & 0.761 & 0.786 & 0.779 & 0.829 & 0.825 & 0.840 & 0.832 & 0.739 & 0.730 & 0.755 & 0.746 \\
\textit{w/o CL}        & 0.826 & 0.817 & 0.851 & 0.845 & 0.785 & 0.776 & 0.796 & 0.789 & 0.835 & 0.827 & 0.849 & 0.843 & 0.753 & 0.749 & 0.764 & 0.755 \\
\textit{w/o PI}        & 0.771 & 0.760 & 0.820 & 0.809 & 0.741 & 0.735 & 0.762 & 0.755 & 0.809 & 0.791 & 0.821 & 0.817 & 0.736 & 0.725 & 0.742 & 0.735 \\ \hline
\end{tabular}%
}

\end{table}

\subsection{Ablation Study}
We design several model variants to investigate the ablation experiments of five crucial components and determine their importance in our model. (I) \textit{Meta-GPS++/SGC:} We use simplifying graph convolution to encode the graph instead of our designed graph encoded layer. (II) \textit{Meta-GPS++/PI:} We exclude the prototype-based initialization parameters and replace them with random initialization parameters. (III) \textit{Meta-GPS++/CL:} We eliminate the adopted contrastive learning module. (IV) \textit{Meta-GPS++/ST}: We delete the designed self-training module. (V) \textit{Meta-GPS++/S$^2$:} We exclude the S$^2$ transformation module, which generates the scaling and shifting vectors, and treat all tasks equally. 

To confirm that concatenation is superior to averaging in learning node embedding, we perform visualization experiments on two selected datasets, \textit{i.e.}, Amazon-Electronics and DBLP. As shown in Fig. \ref{tsne}, our graph layer utilizing the concatenation operator results in improved class separation compared to SGC's use of the average operator. This finding suggests that our graph layer can effectively operate within heterophilic networks. Moreover, we can gain valuable insights based on the results presented in Table \ref{ablation}. First, removing any module leads to a decline in the model's performance in the range of 1-6\%, indicating that all five designed modules play important roles. Second, compared with the original model Meta-GPS++, Meta-GPS++/PI has the largest performance decline. For example, the model's accuracy and F1 score decreased by 4.9\% and 3.8\%, respectively, in the 5-way 5-shot experimental setting on the Amazon-Electronics dataset. This indicates that prototype-based initialization can promote optimal parameter learning and guarantee stable model learning. Third, Meta-GPS++/CL achieves the second-worst performance, and usually leads to a performance drop of around 2\% on all datasets. This indicates that supervised contrastive learning can regulate the distribution of node embeddings, allowing nodes of the same category to cluster together and thus alleviate the negative impact brought by task randomness. Fourth, Meta-GPS++/SGC is effective on homophilic networks but struggles in heterophilic networks. For example, in the 5-way 3-shot setting on the Reddit dataset, the accuracy of Meta-GPS++/SGC decreases by 1.4\%, while in the 5-way 3-shot setting on the Amazon-Electronics dataset, its accuracy decreased by 3.7\%. Because it cannot generate high-quality node representations due to the mixture of ego- and neighbor-embeddings. Fifth, self-training provides rich self-supervision information and task-relevant information, which encourages the model to use the available data more efficiently and alleviates the negative impact of the node distribution bias caused by the limited labeled nodes in the support set. It usually yields a gain of around 1\%. Finally, the $S^2$ module can customize parameters for each task, thus improving the model's generalization ability. Removing it results in a significant performance drop, for example, a decrease of 2.6\% in the 5-way 3-shot setting on the Amazon-Clothing dataset.
%we can obtain the following insights and analysis by observing the results presented in Fig. \ref{ablation}. First, when we eliminate any of the modules, the model's performance receives a different degree of degradation, indicating that all these parts play a critical role in the model. Second, Meta-GPS-SGC performs comparably in homophilic networks but poorly in heterophilic networks. One reasonable explanation is that Meta-GPS-SGC cannot learn powerful node representations from heterophilic networks due to mixing the ego- and neighbor- embeddings. 
%Third, Meta-GPS-PI obtains the worst performance among all model variants, illustrating that adopting prototype-based initialization facilitates the model to learn the optimal parameters and provides guarantees for stable model learning. Finally, customizing the weight parameters for each task is beneficial to improve the generalization of the model.

\begin{figure}
    \centering
    \subfigure[Meta-GPS++ on Reddit]{\includegraphics[width=0.3\textwidth]{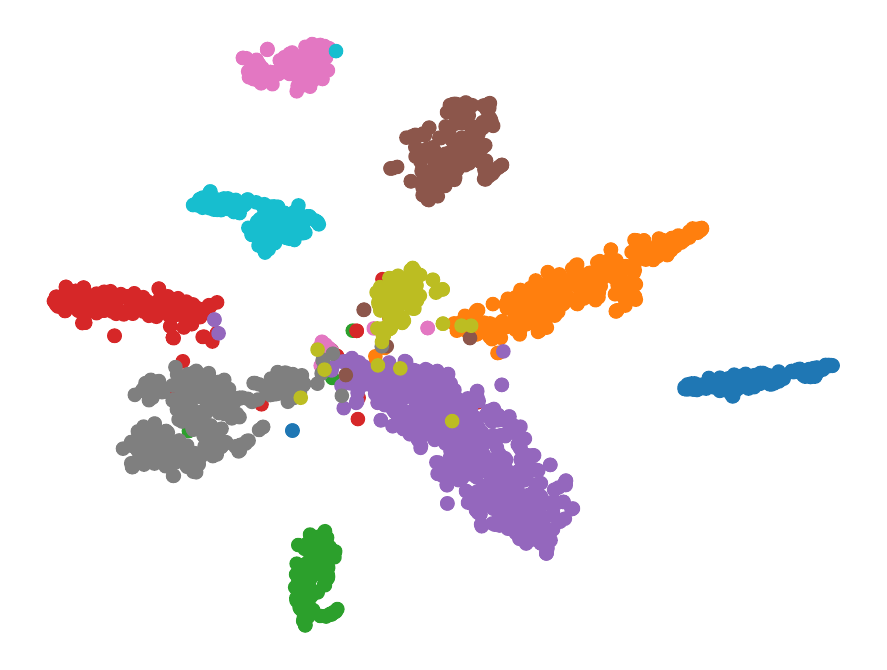}}
    \subfigure[IA-FSNC on Reddit]{\includegraphics[width=0.3\textwidth]{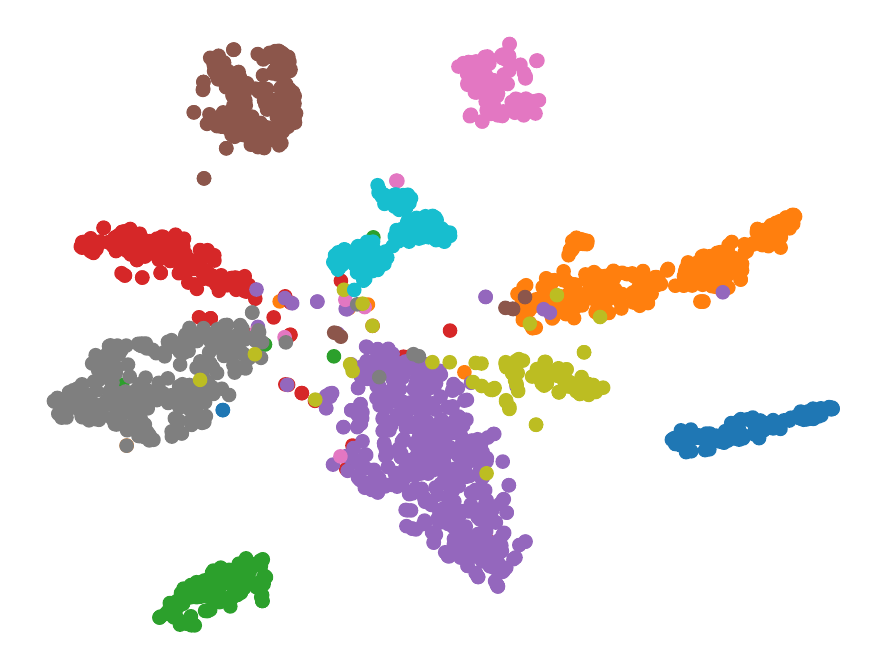}}
    \subfigure[GPN on Reddit]{\includegraphics[width=0.3\textwidth]{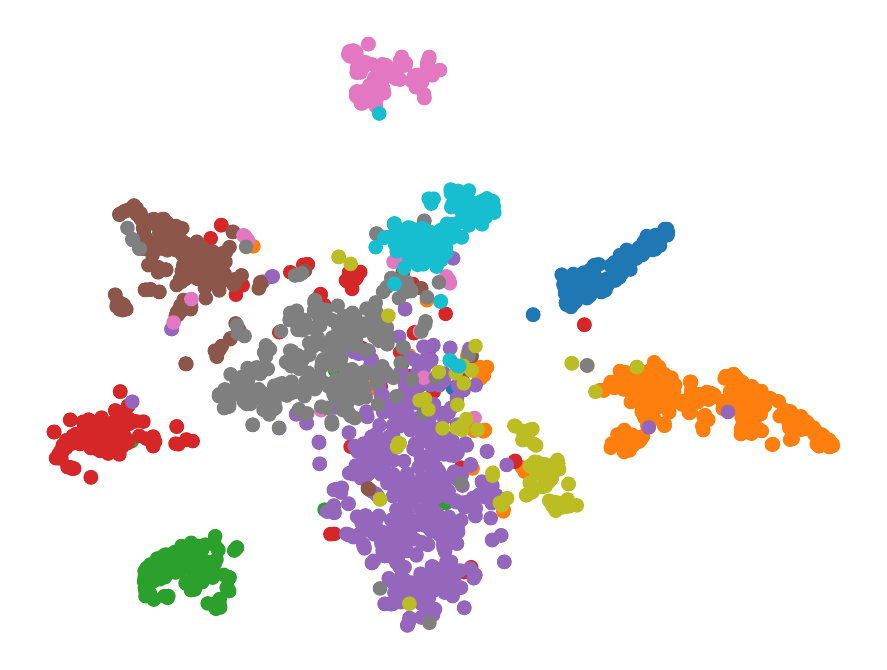}}
    
    \subfigure[Meta-GPS++ on ogbn-products]{\includegraphics[width=0.3\textwidth]{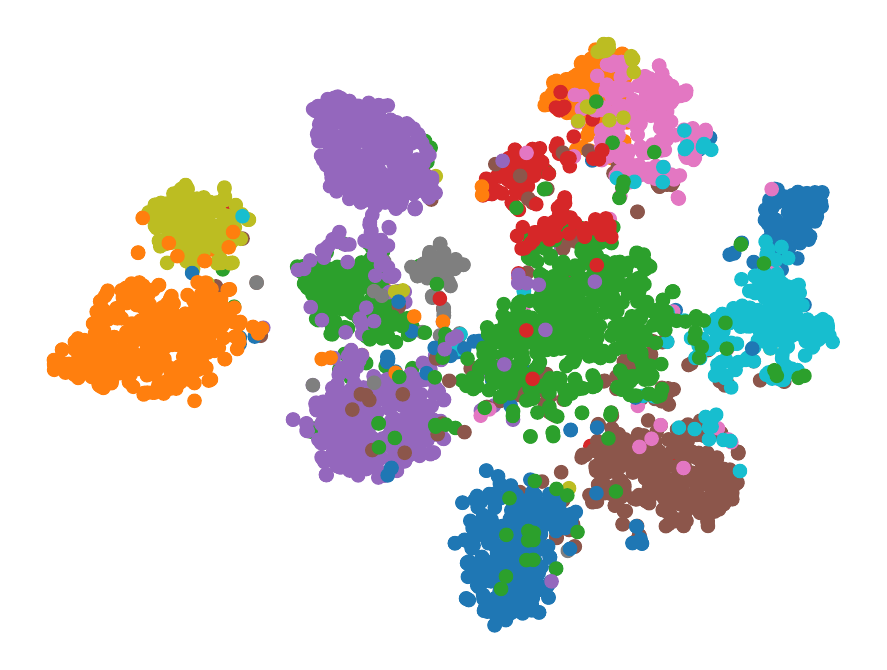}}
    \subfigure[IA-FSNC on ogbn-products]{\includegraphics[width=0.3\textwidth]{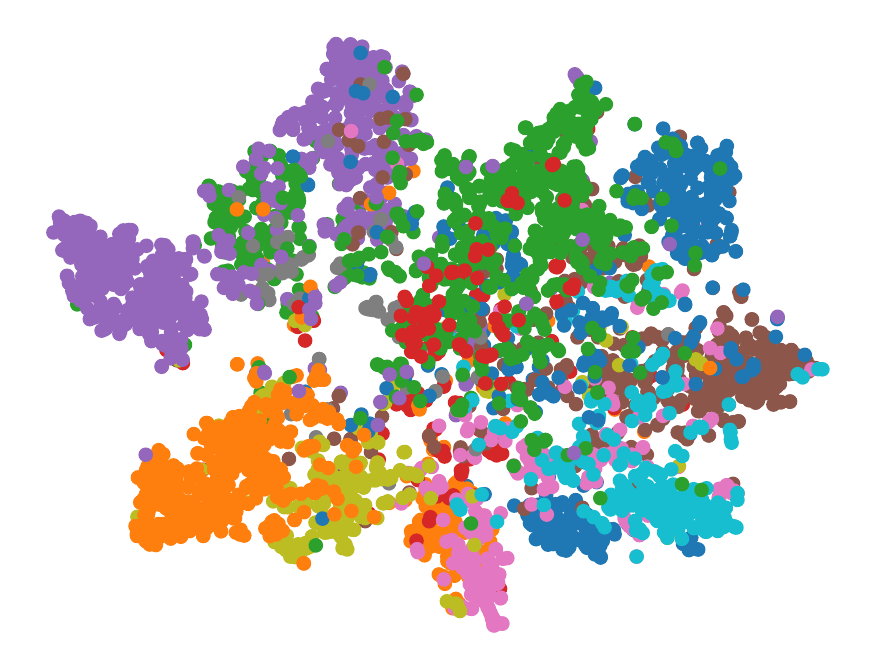}}
    \subfigure[GPN on ogbn-products]{\includegraphics[width=0.3\textwidth]{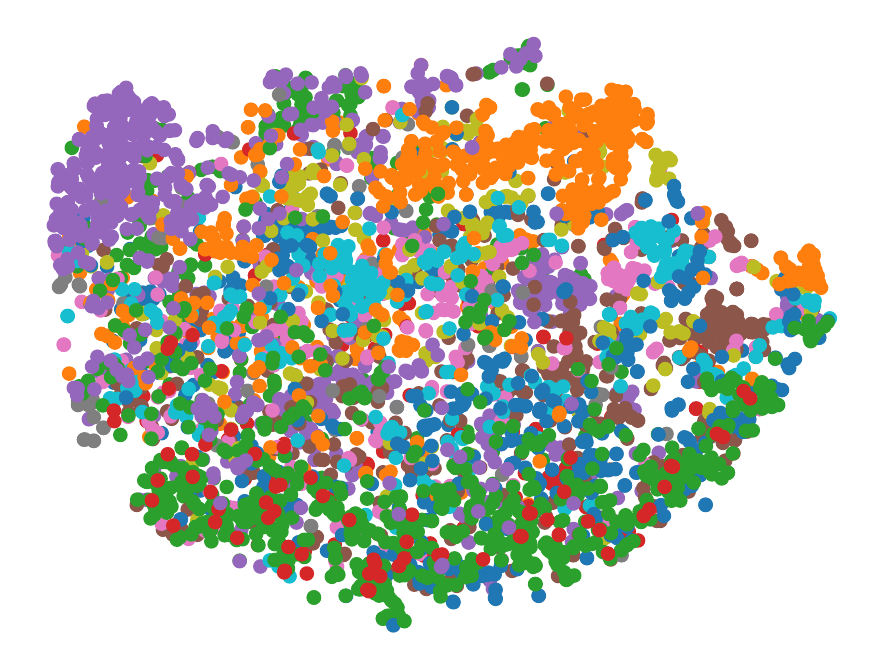}}

    \subfigure[Meta-GPS++ on Amazon-Clothing]{\includegraphics[width=0.3\textwidth]{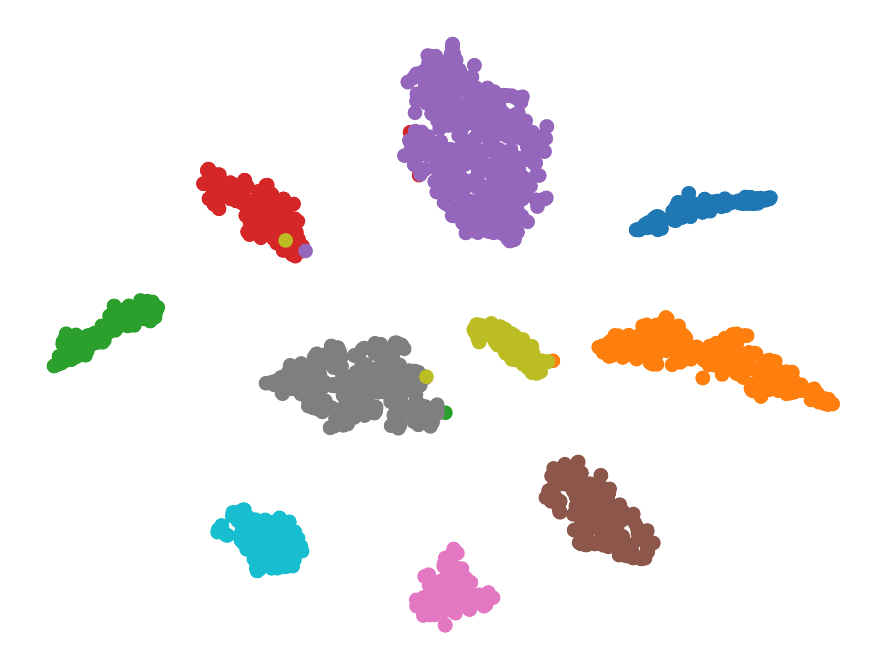}}
    \subfigure[IA-FSNC on Amazon-Clothing]{\includegraphics[width=0.3\textwidth]{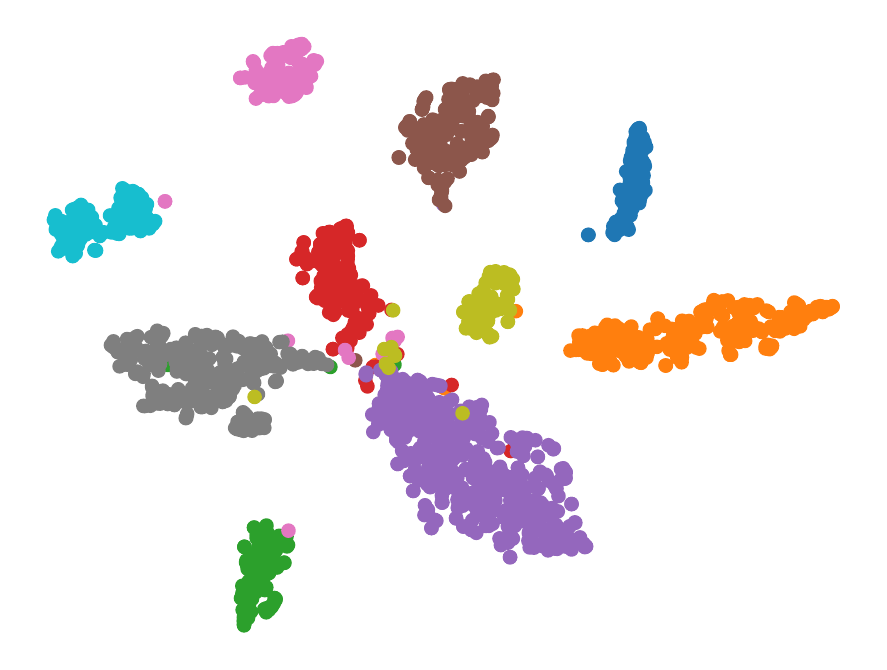}}
    \subfigure[GPN on Amazon-Clothing]{\includegraphics[width=0.3\textwidth]{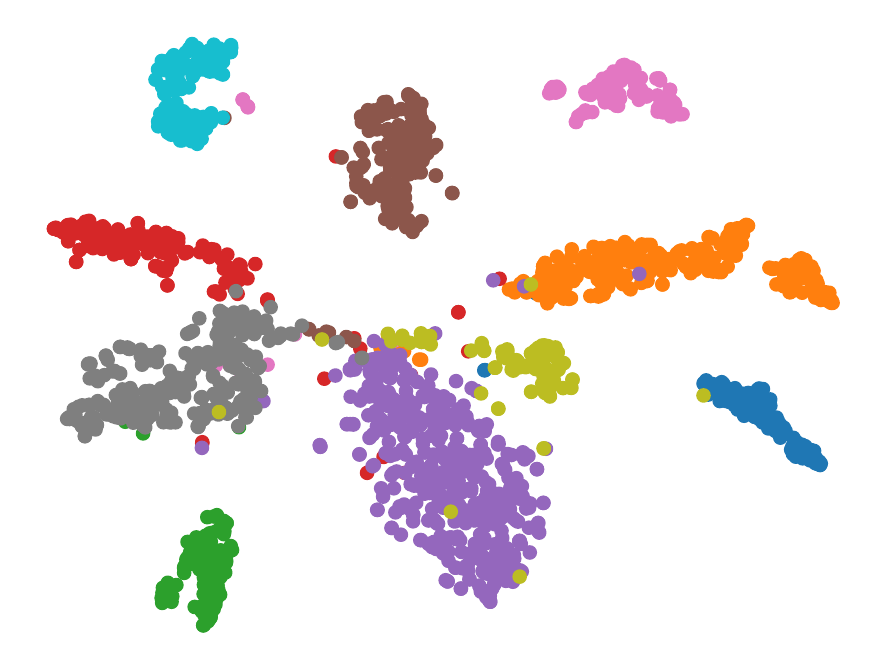}}

    \subfigure[Meta-GPS++ on Cora-Full]{\includegraphics[width=0.3\textwidth]{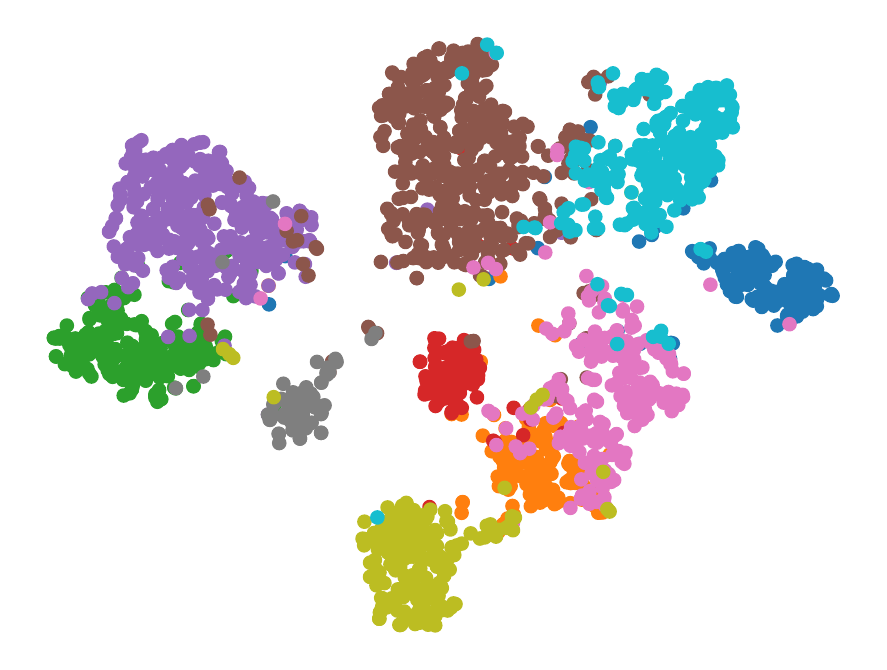}}
    \subfigure[IA-FSNC on Cora-Full]{\includegraphics[width=0.3\textwidth]{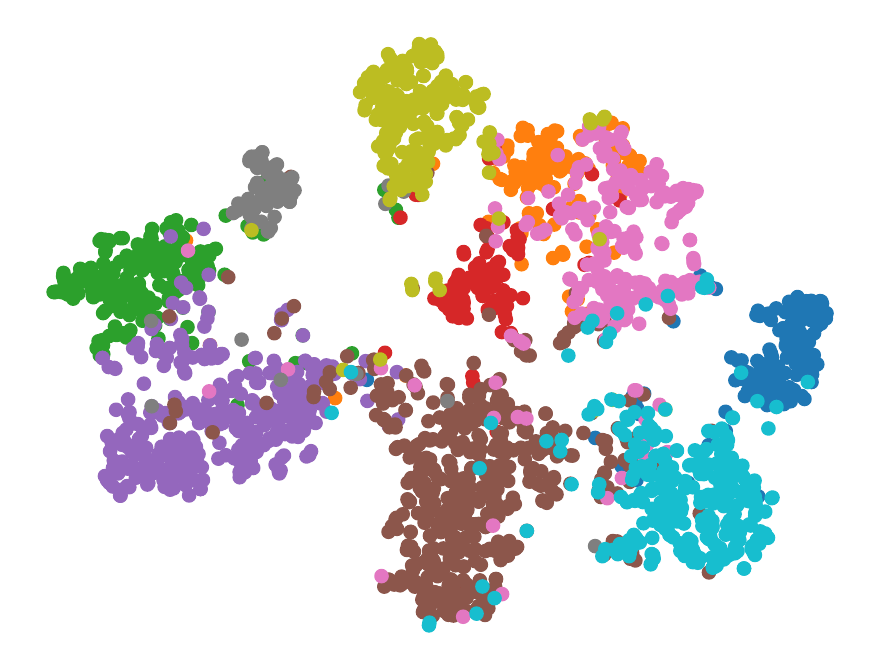}}
    \subfigure[GPN on Cora-Full]{\includegraphics[width=0.3\textwidth]{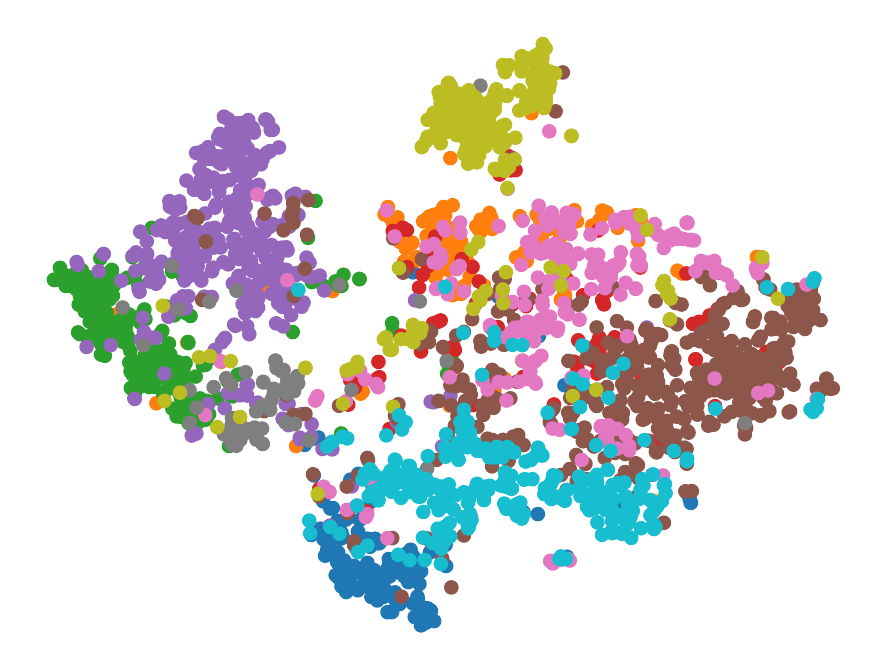}}

    \caption{Visualizations of the learned embeddings of different methods with the partial test sets on several datasets.}
    \label{vis}
\end{figure}
\subsection{Model Visualization}
To further demonstrate the effectiveness of our proposed model, we first visualize the node embeddings learned by Meta-GPS++, as well as two other representative models (\textit{i.e.}, GPN and IA-FSNC), on partial classes from test sets of four chosen datasets in a qualitative way. As shown in Fig. \ref{vis}, Meta-GPS++ can learn more discriminative node features than other models. This makes the node embeddings of the same class closer and those of different classes further away, with clearer class boundaries. Even on challenging datasets like Reddit and ogbn-products, Meta-GPS++ can more clearly observe the clustering phenomenon of node embeddings. These empirical results validate the superiority of our model in learning high-quality node embeddings.

Subsequently, we further investigate the discriminability of the model's learned embeddings in a quantitative way. Here, we adopt two widely used metrics, namely the Silhouette Coefficient (SC) \cite{rousseeuw1987silhouettes} and Davies-Bouldin (DB) index \cite{davies1979cluster}. Here, SC is calculated by the average intra-cluster distance ($a_i$) and the average nearest cluster distance ($b_i$) for each sample, with values ranging from -1 to 1. Higher values indicate better results. It can be defined as follows:
\begin{equation}
    \begin{aligned}
        sc_i = \frac{b_i-a_i}{\max(a_i,b_i)}, \quad
        \text{SC} = \frac{1}{n}\sum\nolimits_{i=1}^{n}sc_i
    \end{aligned}
\end{equation}
where $sc_i$ is the SC value of the sample $i$.

Moreover, DB is computed by the average similarity of each cluster to its most similar cluster, where the similarity is the ratio of the intra-cluster distance to the inter-cluster distance. It measures intra-class variation, with lower values indicating a better data clustering structure. The mathematical expression can be denoted as:
\begin{equation}
    \text{DB}=\frac{1}{M}\sum\nolimits_{i,j=1}^M\max_{i\neq j}\frac{s_i+s_j}{d_{ij}}
\end{equation}
where $M$ is the number of classes, $s_i$ represents the average distance of all data in the $i$-th cluster to the cluster center, and $d_{ij}$ represents the distance between the two cluster centers.

\begin{table}[ht]
\centering
\caption{Qualitative results of the learned embeddings of different methods with the partial test sets on several datasets.}
\label{vis_qualitative}
%\resizebox{0.9\textwidth}{!}{%
\begin{tabular}{c|cccccccc}
\hline
\multirow{2}{*}{Models} & \multicolumn{2}{c}{Reddit} & \multicolumn{2}{c}{ogbn-products} & \multicolumn{2}{c}{Amazon-Clothing} & \multicolumn{2}{c}{Cora-Full} \\ \cline{2-9} 
        & SC     & DB    & SC     & DB    & SC    & DB    & SC    & DB    \\ \hline
GPN     & 0.162 & 1.459 & -0.072 & 4.414 & 0.291 & 1.126 & 0.051 & 2.110 \\
IA-FSNC & 0.270 & 1.145 & 0.026  & 2.057 & 0.365 & 0.979 & 0.211 & 1.345 \\
Ours    & 0.328  & 1.032 & 0.183  & 1.484 & 0.483 & 0.784 & 0.262 & 1.193 \\ \hline
\end{tabular}%
%}
\end{table}

From the qualitative results presented on Table \ref{vis_qualitative}, we clearly observe that our model has higher SC values and lower DB values than other representative models. This indicates that nodes of the same class are more likely to cluster in our model, resulting in clearer class boundaries.
\section{Conclusion}
In this work, we design a novel graph meta-learning model called Meta-GPS++ for few-shot node classification tasks. To achieve this, we make five essential improvements, namely, a graph network encoder, prototype-based parameter initialization, contrastive learning for task randomness, self-training for model regularization, and S$^2$ transformation for suiting different tasks. More specifically, we first adopt an efficient method to learn discriminative node representations on homophilic and heterophilic graphs. Then, we leverage a prototype-based approach to initialize parameters in meta-learning and contrastive learning for regularizing the distribution of node embeddings. Moreover, we also apply self-training to extract valuable information from unlabeled nodes to facilitate model training, in which an S$^2$ (scaling \& shifting) transformation is adopted to learn transferable knowledge from diverse tasks, which can effectively solve the limitations of current models. We conduct thorough experiments under different experimental settings on six real-world datasets. The experimental results empirically show that Meta-GPS++ has significantly superior performance compared to other state-of-the-art models.

%%
%% The acknowledgments section is defined using the "acks" environment
%% (and NOT an unnumbered section). This ensures the proper
%% identification of the section in the article metadata, and the
%% consistent spelling of the heading.
\begin{acks}
Our work is supported by the National Key Research and Development Program of China (No. 2021YFF1201200), the National Natural Science Foundation of China (No. 62172187, No. 62372494, and No. 62372209), Guangdong Universities’ Innovation Team Project (No. 2021KCXTD015), Guangdong Key Disciplines Project (No. 2021ZDJS138 and No. 2022ZDJS139). Fausto Giunchiglia’s work is funded by European Union’s Horizon 2020 FET Proactive Project (No. 823783).
\end{acks}

%%
%% The next two lines define the bibliography style to be used, and
%% the bibliography file.
\bibliographystyle{ACM-Reference-Format}
\bibliography{sample-base}

%%
%% If your work has an appendix, this is the place to put it.
%\appendix

\end{document}